\documentclass[journal,twoside]{IEEEtran}
\newcommand{\ignore}[1]{}

\usepackage[utf8]{inputenc}
\usepackage[T1]{fontenc}
\usepackage{amsmath,epsfig}
\usepackage{multirow}
\usepackage{url}
\usepackage[colorlinks, citecolor=blue, bookmarks]{hyperref}
\usepackage{graphicx}
\usepackage{bm}
\usepackage{amsmath}
\usepackage{amssymb}
\usepackage{cite}
\usepackage{array}
\usepackage{color}
\usepackage{booktabs}
\usepackage{cases}
\usepackage{bm}
\usepackage{fixltx2e} 
\usepackage{makecell} 
\usepackage{epstopdf}
\usepackage{epsfig}

\newcommand{\first}[1]{\textbf{\textcolor[rgb]{1,0,0}{#1}}}
\newcommand{\second}[1]{\textbf{\textcolor[rgb]{0,0,1}{#1}}}
\newcommand{\third}[1]{\textbf{\textcolor[rgb]{0,1,0}{#1}}}

\begin{document}
%
\title{Dual Deep Network for Visual Tracking}

\author{Zhizhen Chi, Hongyang Li, \emph{Student Member, IEEE}, Huchuan Lu, \emph{Senior Member, IEEE}, \\
and Ming-Hsuan Yang, \emph{Senior Member, IEEE}

\thanks{Z. Chi and H. Lu are with School of Information and Communication
	Engineering, Dalian University of Technology, Dalian 116024, China.
	Email: \texttt{zhizhenchi@gmail.com}, \texttt{lhuchuan@dlut.edu.cn}.}

\thanks{H. Li is 
	with Department of Electronic
	Engineering, The Chinese University of Hong Kong,
	Hong Kong.
	Email: \texttt{yangli@ee.cuhk.edu.hk}.}

\thanks{M. Yang is with University of California at Merced,
	Merced, CA 95344, USA.
	Email: \texttt{mhyang@ucmerced.edu}.}

}

\markboth{Submitted to IEEE Transactions on Image Processing.}
{Chi, \MakeLowercase{\textit{et al.}:} Dual Deep Network for Visual Tracking.}

\maketitle

\begin{abstract}
Visual tracking addresses the problem of identifying and localizing an unknown target in a video given the target specified by a bounding box in the first frame.
In this paper, we propose a dual network to better utilize features among layers for visual tracking.
%
It is observed that features in higher layers encode  semantic context while its counterparts in lower layers are  sensitive to discriminative appearance.
Thus we exploit the hierarchical features in  different layers of a deep model 
and design a dual  structure to obtain better  feature representation from various streams, which is rarely investigated in previous work.
To highlight geometric contours of the target, we integrate the hierarchical feature maps with an edge detector as the coarse prior maps to further embed local details around the target.
To leverage the robustness of our dual network, we train it with random patches  measuring the similarities between the network activation and target appearance, which serves as a regularization to enforce the dual network to focus on target object.
The proposed dual network is updated online in a unique manner based on the observation that the target being tracked in consecutive frames should share more similar feature representations than those in the surrounding background.
It is also found that for a target object, the prior maps can help further enhance performance by passing message into the output maps of the dual network.
Therefore, an independent component analysis with reference algorithm (ICA-R) is employed to extract 
target context using prior maps as guidance.
Online tracking is conducted by maximizing the posterior estimate on the final maps with stochastic and periodic update.
Quantitative and qualitative evaluations on two large-scale benchmark data sets show that the proposed algorithm
performs favourably against the state-of-the-arts.

\end{abstract}

\begin{IEEEkeywords}
Visual Tracking, Deep Neural Network, Independent Component Analysis with Reference.
\end{IEEEkeywords}

\IEEEpeerreviewmaketitle

\section{Introduction}
\label{intro}

Visual tracking has long been a fundamental research problem in computer
vision~\cite{ren07,ross2008,grabner08,mei09,hare2011,jia2012visual,zhong14,zhang2014meem}.
In a typical
scenario, an unknown target specified by a bounding box in the first frame is to be tracked in the subsequent
frames. Although numerous algorithms have been proposed for visual tracking, it remains a challenging problem
for a tracker to handle large appearance variations, abrupt motion, severe occlusions, and background
clutters.
To address these challenges, recent approaches resort to robust representations such as Haar-like
features~\cite{hare2011}, histograms~\cite{jia2012visual}, and sparse features~\cite{zhong14}.
%
However, these low-level hand-crafted features are not effective in capturing semantic information of targets due
to limited discriminative strength of representation schemes.

\begin{figure}[t]
	\centering
	\includegraphics[width=.47\textwidth]{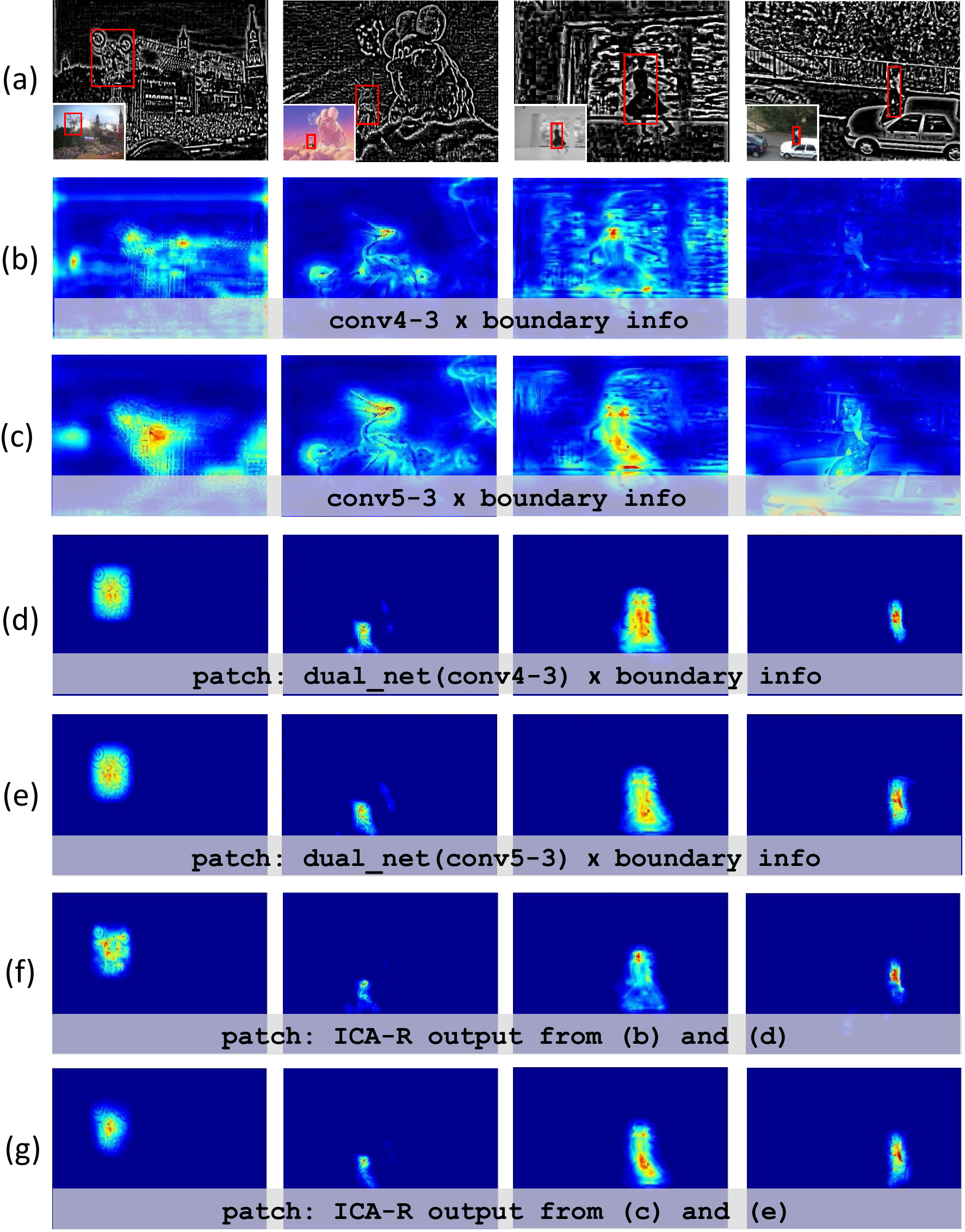}
	\vspace{-2mm}
	\caption{Feature map breakdown of different stages.
		(a) Boundary maps produced by LoG filters.
		The insets show the tracking frames.
		(b)-(c) Coarse prior maps
		from the feature maps in VGG \texttt{conv4-3} and \texttt{conv5-3} layers 
		and the boundary maps.
		%
		(d)-(e) Integrated maps 
		of the outputs of the dual network
		multiplied by the boundary maps.
		(f)-(g) ICA-R results
		which combine the mixed signals
		(d)(e)
		with reference
		(b)(c).
			Note that maps from (d) to (g) are visualizations of the tracked targets and resized only for better illustration.
	}
	\label{fig:figure2}
	\vspace{-0.4cm}
\end{figure}

Recent years have witnessed the resurgence of visual representation learning from raw pixels using deep convolutional neural networks (ConvNet) \cite{krizhevsky12,li2016multi}.
Driven by the large-scale visual data sets and fast development of computation resources,
methods based on deep models have achieved state-of-the-art results in a wide range of
vision tasks, such as image classification~\cite{krizhevsky12,simonyan2014very,he2015deep}, object detection~\cite{girshick14,girshick2015fast},
saliency detection \cite{li2015inner,li2016cnn_sal}, etc.
%
%
%
For visual tracking, one bottleneck of applying ConvNet is the lack of training data for a sequence as typically only one labelled bounding box in the first frame is given.
Although this problem may be alleviated by using more data during online update to learn and refine a generic object model,
it entails a significant increase in computational complexity.
An alternative is to transfer the ConvNet features learned offline to online tracking \cite{wang2012,wang13,hong2015tracking}.
However, simply regarding the deep model as a black box feature extractor
may not complement offline training with online tracking, and previous work \cite{li_deeptrack,wang2015video,nam2015learning} do not distinguish or utilize various  information carried among different levels of layers.
%
%

We argue that
 features at different levels of a ConvNet capture image properties
 and should be fully utilized in a proper manner for visual tracking.
It is observed that,
on one hand,
features at lower layers encode more fine-grained details to locate target objects precisely before being pooled or stridden into deeper layers
(see Figure \ref{fig:figure2}(b));
on the other hand,
features at higher layers capture more high-level semantics useful for
discriminating objects from the cluttered background \cite{bappy2016cnn} and are robust to deformation and occlusion (see Figure \ref{fig:figure2}(c)).
%
%
%
%
These observations motivate us to consider multiple layers of a ConvNet into
tracking to better obtain task-specific target features.
One may resort to a hypercolumn \cite{hypercolumn} framework to combine feature maps of all layers.
In fact, two layers are complementary enough for different hierarchical features. Stacking more or making net deeper may boost performance\footnote{In fact, we verify in preliminary experiments that simplying stacking the feature maps of three layers does not perform well than that of using two layers alone. See Table~\ref{ablation2} Section~\ref{ablationstudy} for details.}, but how to combine features from more cascaded layers 
are beyond the work of this paper and need more time to tune (more parameters and overfitting problem).
As seen in Figure \ref{fig:figure2}(d)-(e), features in the patch after two streams of a dual network focus on the target being tracked and they can be further refined with
prior feature maps from VGG model in an ICA with reference manner \cite{LuR06} (see Figure \ref{fig:figure2}(f)-(g)).
The final estimation of a target location takes advantages of both high-level semantics and low-level details from the two streams.



\begin{figure}
  \centering
   \includegraphics[width=.48\textwidth]{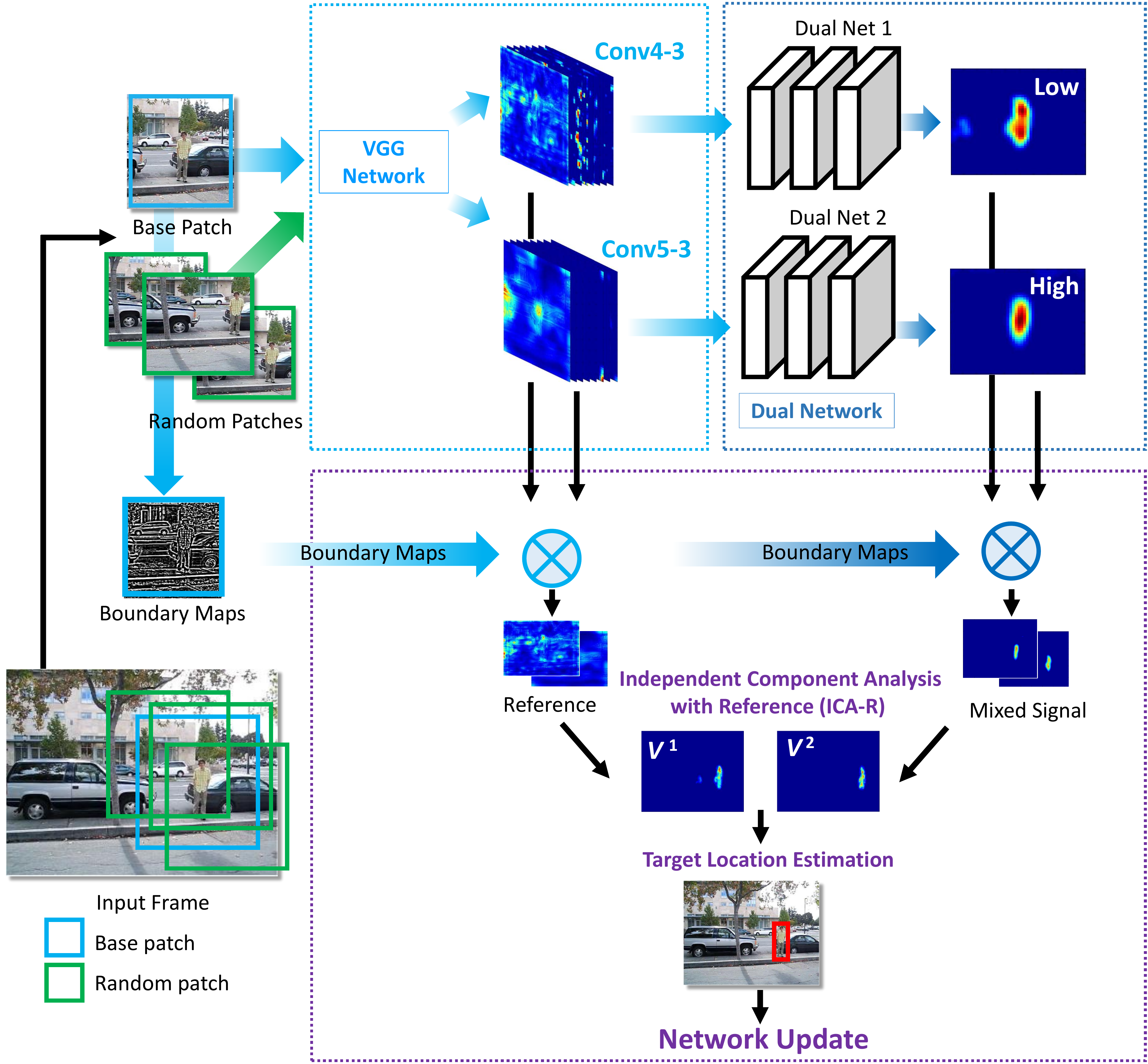}
  \caption{Pipeline of our algorithm.
  For each frame, a base patch as well as a bunch of random patches are generated and fed into the dual network for training. For brevity, we only show the workflow of a base patch (random patches, marked as green, also follow the same procedure).
The output feature maps of the VGG and the dual network are combined with boundary maps to generate a set of robust feature representations, which are known as reference and mixed signal, respectively.
ICA-R outputs from two streams are combined to estimate the target location and  further used for online tracking where  drifts and occlusion issues can be handled.
  }
  \label{pipeline}
\end{figure}

In this paper,
we propose a dual network based tracker (DNT) in a self-supervised learning framework,
which incorporates both high-level semantic context and low-level spatial structure from different layers (see Figure~\ref{pipeline}).
%
%
First, we extract the deep features, combine them with the low-level boundary maps derived from the Laplacian of Gaussian (LoG) edge detector,
and incorporate these features to obtain a more concise feature map via the independent component analysis with reference (ICA-R) method \cite{LuR06}
(Section~\ref{subsec:overview}).
Second,
the dual network training is detailed
where center-shifted random patches 
are generated to augment the number of positive training samples (Section~\ref{subsec:network}).
Third,
the tracking framework with deep features is embedded into the motion and observation model with a
module to detect drift and occlusion (Section~\ref{subsec:track}).
The dual network is updated online with both stochastic and periodic schemes to address the drift and occlusion problems (Section~\ref{sec:update}).
The  implementation of our proposed DNT algorithm is available at
\href{https://github.com/chizhizhen/}{\texttt{ https://github.com/chizhizhen/DNT}}.

\smallskip
The contributions of this work are summarized as follows:
\vspace{-.1cm}
\begin{itemize}
	\item 
	Boundary and shape information are embedded within the deep model framework.
	Low level texture features generated by an edge detector are incorporated into the hierarchical deep features to obtain a coarse prior map.
	The salient target contour is improved via the ICA-R method for a better observation likelihood model.	
	\smallskip	
  \item A dual network, which extracts representative features from two distinct layers, is proposed.
  Location translation is exploited as data augmentation to enable the dual network for robustly discriminating foreground target from background clutters.
  \smallskip
  \item Both stochastic and periodic update mechanisms are presented to handle occlusion or drift during online tracking and to alleviate
issues with noisy updates caused by inaccurate training samples.
\end{itemize}

The rest of the paper is organised as follows. In Section~\ref{sec:related-work}, we first review related work and discuss our improvements. The feature extraction, online tracking and dual network training and updating are analysed in Section~\ref{sec:dualnet}. We evaluate the proposed tracker on two large-scale benchmark data sets~\cite{wu2013online,kristan2015visual,wu2015object} in Section~\ref{sec:experiment} and substantial performance improvement is achieved
over the state-of-the-art methods.
Furthermore, we conduct a series of ablation studies to demonstrate the effectiveness of each component.
%

\section{Related work}\label{sec:related-work}
Most tracking methods can be categorized into either generative or discriminative model based trackers.
In generative models, candidate regions are searched to select the most probable one as the tracking result.
Some representative methods are based on principal component analysis~\cite{ross2008}~to model target appearance.
Sparse coding~\cite{mei09,zhong14} is also introduced
using different dictionaries or regulation terms.
A target object is then reconstructed by a sparse linear combination of target and trivial templates.
%
In discriminative models, tracking is considered as a binary classification problem to
separate the target from background.
Avidan~\cite{avidan2004support}~introduces an optical flow approach with a SVM classifier for visual tracking.
The multiple instance learning~\cite{babenko11} and structured output SVM~\cite{hare2011}~are also applied to visual tracking.
Recently, correlation filters~\cite{CSK12,bolme2010visual,danelljan2014accurate,henriques15} have been used for efficient and effective visual tracking.
%
%
In addition, subsequent algorithms based on correlation filters
with hand-crafted features (such as HOG features~\cite{henriques15}~or color-attributes~\cite{danelljan2014adaptive})~considerably improve tracking accuracy.

%

Hierarchical feature representations from deep learning have been shown to be effective for visual tracking~\cite{wang13,nam2015learning,li_deeptrack,wangvisual,machao15,hong2015tracking,ondruska2016deep,cui2016recurrently,wang2015video}.
%
%
%
Wang and Yeung~\cite{wang13} propose a deep learning approach using a stacked denoising autoencoder where the model is
trained in an unsupervised fashion on an auxiliary data set
and directly used for online tracking.
On one hand, image samples used for pretraining are cropped from frames in a downsampling scanning approach, which leads to significant
amount of redundant computation.
On the other hand, features learned offline from an auxiliary data set may not adapt well to specific sequences containing targets for visual tracking.
The DeepTrack~\cite{li_deeptrack} method treats tracking as foreground-background classification task with a ConvNet trained online.
Similarly, an online SVM with a pretrained CNN is presented for visual tracking in~\cite{hong2015tracking}, and exploits multiple levels of representations of CNN by back-propagating target information through the whole network until the image domain.
Such schemes to directly learn two-layer or SVM classifiers  require a large number of positive and negative samples, which is impractical for online tracking.
%
%
Ma and Wang~\emph{et al.}~\cite{machao15,wangvisual} analyze different characteristics of each convolutional layer from which features are combined to represent target
objects.
In~\cite{machao15}, feature maps derived from the offline model are directly used to learn the Kernelized Correlation Filter
without online updating network.
In~\cite{wangvisual}, two online networks with shallow architectures are trained and updated for feature selection and representation,
where the models can remove unreliable and noisy feature maps to alleviate the overfitting problem.
A multi-directional recurrent neural network  is employed in the RTT tracker \cite{cui2016recurrently} to capture long-range contextual cues by traversing a candidate spatial region from multiple directions.

In this paper, we have tackled three main issues with the existing deep model based trackers.
First, the features used for tracking are learned from a pretrained CNN on other data sets. It is obvious that they may not be adaptive to target appearance in visual tracking. Thus, we train an adaptation module online for each specific sequence.
Second, texture details of the target objects are not explicitly considered. In the papers~\cite{wangvisual,machao15}, after feature maps of ConvNets are pooled or convoluted with consecutive strides, fewer texture details are preserved and only salient object responses are used for tracking.
Hand-crafted boundary and shape information can contribute greatly to accurate object localization
and can be integrated with the deep representations to generate better candidates during visual tracking.
Third, surrounding background and similar objects around the target objects are not fully exploited for learning more effective representations.
Different patch translation schemes can be used as the data augmentation for online training and model update.
We address these issues in the proposed algorithm.

\section{Dual Network Tracker}\label{sec:dualnet}
\label{sec:blind}

\subsection{Deep Feature Extraction with ICA-R}
\label{subsec:overview}
We use the 16-layer VGG model~\cite{simonyan2014very} pretrained on the ImageNet data set
as the feature representation, where the network contains 13 convolutional layers and 3 fully-connected layers.
Specifically, the 10-th and 13-th convolutional layer are used to extract hierarchical features, denoted as \texttt{conv4-3} and \texttt{conv5-3}, respectively.
These ConvNet features are integrated with boundary and shape information during tracking, as well as fed into the dual network for deeper representations.

Given an input frame $I$, we convolve it with the Laplacian of Gaussian filter \cite{marr1980theory} and output the boundary map as $\textbf{f}_{E}$.
Figure \ref{fig:figure2}(a) shows the boundary maps of four sample frames.
%
Let $\textbf{g}_V^k(x,y), \textbf{g}_D^k(x,y)$
denote the output ConvNet feature maps generated by the VGG network and the dual network respectively, where $k= \{1,2\}$ indicates the layer index
\texttt{conv4-3} or \texttt{conv5-3} and $x, y$ denote the spatial coordinates.
The integrated feature maps $\textbf{h}^{k}_{V}, \textbf{h}^{k}_{D}$, which take both semantic and boundary information into consideration, are defined as follows:
\begin{align} \label{integration}
\textbf{g}_D^k &= \texttt{dual\_net}(\textbf{g}_V^k),\\
\textbf{h}^{k}_{V}(x,y)&=\textbf{f}_E(x,y) \odot \textbf{g}_V^k(x,y), \\
\textbf{h}^{k}_{D}(x,y)&=\textbf{f}_E(x,y) \odot \textbf{g}_D^k(x,y),
\end{align}
%
where $\odot$ denotes the element-wise multiplication, and $\mathtt{dual\_net}(\cdot)$ represents the forward passing from the VGG output as the input to the dual network.
These integrated feature maps provide the boundary information as a guidance for target localization and serve as the coarse prior maps used for online tracking.
Figure \ref{fig:figure2}(b)-(c) depicts the integrated prior maps $\textbf{h}_V^k$ which
combine the deep semantic features and low-level texture information. Figure \ref{fig:figure2}(d)-(e)
shows the integrated maps $\textbf{h}_D^k$ after the dual network multiplied by the boundary information to have more accurate target locations.

The independent component analysis with reference (ICA-R) method~\cite{LuR06} is
developed to extract the desired signal among source components guided by references.
Given the prior maps $\textbf{h}_V^k$ as reference and $\textbf{h}_D^k$ as mixed components (signals),
we seek a projection space $s=\textbf{w}_{I}^{T} \textbf{h}_D^k$ to separate the desired signal.
For the tracking task, we derive the ICA-R results $\textbf{v}^k(x,y)$ with a fast one-unit ICA-R algorithm.
Its core idea is to maximize the negentropy $J(s)$ given by
\begin{gather}\label{ica}
J(s) \approx \rho \big \| \mathbb{E}[ \mathcal{Q}(s) ] - \mathbb{E}[ \mathcal{Q}(\epsilon) ]   \big \| ^2\\
\text{s.t.}~~~ \varepsilon(s,\textbf{h}_V)  \leq \xi\nonumber
\end{gather}
where $\mathcal{Q}(x)= \log \cosh(x)$ is the non-quadratic function, $\rho$ is a positive constant,
$\epsilon$ is a Gaussian variable with zero mean and unit variance,
 $\varepsilon(\cdot)$ is a norm function and $\mathbb{E}[x]$ is the expectation of $x$.
For brevity, we drop the superscript of $k$ in the notations above.
The closeness measure $\varepsilon(s,\textbf{h}_V)$ is defined to achieve its minimum when $\textbf{w}=\textbf{w}^{opt}$ and $\xi$ is a threshold.
To find the optical $\textbf{w}^{opt}$, a Newton-like learning method is adopted to iteratively update the vector \textbf{w} and details can be obtained through the Lagrangian duality theory
by introducing a slack variable and regularization term in~\cite{LuR06}.

The motivation for using ICA-R is to describe the \textit{shape} information of a target object more clearly and robustly after extracting the deep features.
Given Gaussian maps as labels, we find in some cases, the output cannot describe accurate contour of target. Thus we associate ICA with net outputs to compensate for shape info. Given $h_V$ as a reference to $h_D$, Gaussian shape is transformed to accurate target shape.
With boundary information considered in $\textbf{f}_E$, the low-level texture information (shape and boundary) and high-level semantic context are fully exploited in the feature space.
Figure \ref{fig:figure2}(f)-(g) shows the ICA-R maps $\textbf{v}^k(x,y)$, which
will be fed into the observation model in the Bayesian framework for visual tracking.

\subsection{Dual Deep Network}
\label{subsec:network}
We formulate a \textit{dual}~(two different functionalities) deep network, which contains two networks with the same structure but different weights learned during training, as an adaptation module to transfer features learned offline to specific target objects.
The dual network consists of three convolutional layers followed by a nonlinear activation.
These three convolutional layers have 128, 64, 64 convolutional kernels of size $7\times7,   ~ 5\times5,  ~   3\times3$, respectively. Note that the output feature map of the last layer is able to highlight salient objects from the input patch.
We train the dual network with fully supervised ground truth in the first frame of each sequence,
and update the model 
according to the similarity of the target appearance in the current and last best tracked frames
 based on the fact that the target in these frames
 should correspond to the same instance of the moving object.
Specifically, given an input frame $I$, we form a patch set $\mathcal{S}_I = \{p, \tilde{p}^{(i)}\}$
that consists of both base and random patches, where $i$ indicates the index of random patches.
A base patch $p$ is centered at the ground truth in the first frame of an image sequence.
A random patch $\tilde{p}^{(i)}$ is a center-shifted version of the base one via location translation and contains more background clutters or similar surrounding objects.
%
Random patches are useful for training in two aspects.
First, they cover more background and distracting objects around the target object.
The salient target object will be well distinguished from background, and the
distracting objects with similar appearance to the target will be suppressed.
Second, considering the scarcity of training data in visual tracking,
these samples serve as data augmentation to facilitate learning the dual network
more effectively.

\begin{figure}[t]
    \centering
	\includegraphics[width=0.49\textwidth,height=0.48\linewidth]{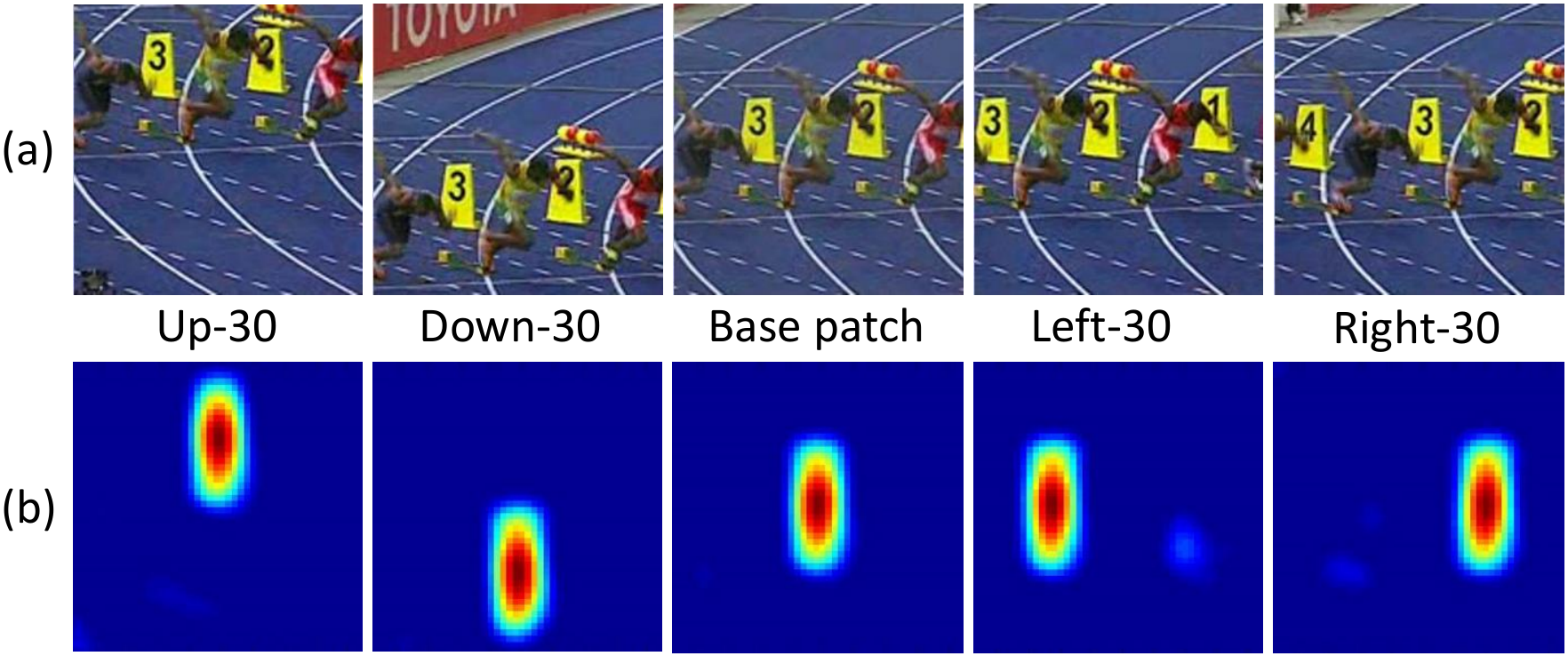}
	\vspace{-6mm}
	\caption{Examples of center-shifted random patches. (a) Both vertical and horizontal patches are randomly  shifted from the base one. Here we use 30 pixels as an illustrative coordinate offset.
		(b) Heat maps generated by the dual network based on the corresponding input patches.
		Here the base patch is the initial search region cropped from the current frame that contains the object of interest.}
	\label{fig:negative_update}
	\vspace{-4mm}
\end{figure}

Figure~\ref{fig:negative_update} illustrates different random patches and their output maps
$\textbf{g}_D$ accordingly.
%
%
When random patches are sampled out of the image boundary, we add zero padding accordingly.
From the examples we observe that the dual network is more robust to discriminate target from background or distractors at different locations.

Let $m$ be the input patch and $\textbf{g}_V^k(m)$ denotes the output of the VGG network,
we obtain the output of the dual network:
\begin{equation}
\textbf{g}_D^k(m) = \texttt{dual\_net}(\textbf{g}_V^k(m)),
\end{equation}
where $k$ is the layer index \texttt{conv4-3} or \texttt{conv5-3} defined earlier.
In the following, we omit the notation $k$ for presentation clarity.
Equipped with the notations aforementioned, we
compute the similarity ranking loss between a base patch and random patch as:
\begin{equation}\label{similarity}
  \ell(p, \tilde{p}^{(i)}) = \mathcal{D} \big( \textbf{g}_D(p), \textbf{t}(p) \big)
  +\mathcal{D} \big( \textbf{g}_D(\tilde{p}^{(i)}), \textbf{t}(\tilde{p}^{(i)}) \big),
\end{equation}
where $\textbf{t}(\cdot)$ denotes a 2-dimensional Gaussian target map generated according to the target location
and $\mathcal{D}(x_1, x_2)$ is the Euclidean distance between two maps.

The loss function enforces that the target object is distinguished from the
surrounding background, and that the distracting objects are suppressed to reduce the risk of drifting towards similar objects around the target.
Note that (\ref{similarity}) resembles in some way the approach in \cite{xiaolong}
where three types of patches are used: original query patch $x$ in the initial frame of tracking, tracked patch $x^+$ in the last frame and random patch $x^-$ from other videos, and which are fed into the network for training.
In \cite{xiaolong}, the model is trained based on a hypothesis that
the distance between the query patch and the tracked
one should be smaller than that between the query patch and the random one.
It is trained in an unsupervised manner for image classification,
which differs from our framework in both configuration and task.
The objective function to initially train the dual network is thereby constructed as follows:
\begin{equation}
\arg \min_\textbf{W}\frac{1}{N}\sum_{i=1}^{N}\ell(\mathit{p},\mathit{\tilde{p}}^{(i)})+\beta{\|\mathbf{W}\|}_2^2,
\label{eq:loss_func}
\end{equation}
where~$\mathbf{W}$~is the weights of the dual network, $\beta$ indicates the weight decay and $\mathit{N}$ represents the number of random patches during one iteration.

%



\subsection{Target location estimation with occlusion and drifts}
\label{subsec:track}
Our tracking algorithm is formulated within the Bayesian framework in which the maximum a posterior estimate is computed based on the likelihood of the candidate
belonging to the target.
Given a new frame, target location is denoted as $\mathit{Z}_t=(x,y,\sigma)$, where $x, y$ and $\sigma$ represent the center coordinates
and scale of the bounding box, respectively. The motion model is assumed to be Gaussian distributed, \textit{i.e.},
$ {\texttt{prob}}(\mathit{Z}_t|\mathit{Z}_{t-1})=\mathcal{N}(\mathit{Z}_t;\mathit{Z}_{t-1},\Psi)$,
where~$\Psi$~is the diagonal covariance matrix that indicates the variance of location and scale parameters.

The sampled candidate regions are normalized to the canonical size maps $\{M^{(r)}_t \}_{r=1}^{R}$
%
with ${v}_t^{(r)}(i,j)$ being the value derived from \eqref{ica} at location $(i,j)$ of the \textit{r}-th candidate at time $t$.
We omit the notation $t$ in the following context for brevity and use it when necessary.
The value in pixel ($r,i,j$) is then normalized as follows:
\begin{equation*}
\widetilde{v}^{(r)}(i,j)=
\begin{cases}
{v}^{(r)}(i,j), & \text{if pixel}~(r,i,j) \in \text{target}, \\
0, & \text{if pixel}  ~(r,i,j) \in \text{background}.
\end{cases}
\end{equation*}
Hence the confidence of the $r$-th candidate is computed as the sum of all the heat map values within the canonical size maps:
$\mathit{C^{(r)}}=\sum_{(i,j)\in{M^{(r)}}}\widetilde{v}^{(r)}(i,j)$.

%
For the tracker to handle large scale variation, we weight $C^{(r)}$ with respect to the size of each candidate:
\vspace{-1mm}
\begin{equation}\label{scale}
\widehat{\mathit{C}}^{(r)}= \frac{ s(Z_t^{(r)}) }{ s(Z^{*}_{t-1}) }
C^{(r)},
\end{equation}
where~$s(\cdot)$~represents the area of the candidate region, $Z_t^{(r)}$ is the $r$-th target location at time $t$ and $Z^{*}_{t-1}$ being the tracking result at the last frame.
%
If the target candidates have higher confidence, the ones with larger area size should be weighted more. If the target candidates are composed of more zeros inside, the ones with larger area size should be weighted less.
This weight scheme facilitates the proposed observation model adaptive to scale variation.

We extend the notation from $\widehat{\mathit{C}}^{(r)}$ to $\widehat{\mathit{C}}^{(r,k)}$
where $k=\{1,2\}$ indicates the confidence of the $r$-th candidate is generated from the ICA-R maps
$\textbf{v}^k$.
%
The final target localization is regarded as a weighted sum from the two hierarchical layers and computed as:
\begin{equation}
\widehat{C}_t^{*}=\arg \max_{r} ~ \lambda \widehat{C}_t^{(r,1)} +
(1-\lambda)\widehat{C}_t^{(r,2)},
\label{maxconfidence}
\end{equation}
where $\lambda$ is a regularization term and $*$ corresponds to the best candidate state $Z_t^{*}$ in the current frame.
The target location is estimated by taking the maximum of~\eqref{maxconfidence} considering both two streams.

\medskip
\noindent
\textbf{Occlusion and Drifts.}
With the proposed observation model based on deep features and ICA-R maps,
we present an efficient method to handle occlusion and drift during tracking.
For a given region candidate $Z_t^{(r)}$ at time  \emph{t}, the corresponding confidence score
$\widehat{C}_t^{(r)}$ is roughly bounded within the range
$\big[0,   s(Z_t^{(r)}) \big]$.
The upper bound indicates that all pixels in the sample region are assigned with the highest confidence to be target object, while the lower bound
indicates all pixels belong to the background.
We set a threshold~$\theta$~to detect object drift or heavy occlusion if the following inequality holds:
\begin{equation}
\mu_C-\frac{\widehat{C}_t^{*}}{s(Z^{*}_{t-1})} < \theta,
\label{detect_occlusion}
\end{equation}
where~$\mu_C$~is the average confidence score of target estimates.
The numerator of the fraction reflects the difference between
the confidence
of the current frame and the average
confidence of the target object.
%
The denominator is a normalization term to
confine the fractional results.
There are two cases where the inequality holds:
\vspace{-.1cm}
\begin{itemize}
\item The numerator is large, which means the confidence $\widehat{C}_t^{*}$ of the current frame is much larger than the average
	confidence.
That means distracting objects similar to the target object appear in proximity, which may lead to the tracker drifting to similar objects.
%
With a proper update mechanism, the proposed tracker is able to recover from drifts.
\item The denominator is small, which means the target object is occluded by unknown objects and only a few pixels are assigned with high values.
%
The occluded parts are weighted less since they are mistaken as background.
Hence we take into consideration the surrounding background while updating the dual network to adapt the appearance model to this occlusion case.
\end{itemize}

\subsection{Online Update}
\label{sec:update}
We update the model features to account for appearance variation, and at
the same time maintain the best tracked target features in the last few frames.
%
%
One key observation in visual tracking is that the feature representations for two best tracked frames should be similar.
Our dual network is updated stochastically and periodically
where the former is to distinguish the target from the background or similar objects, and the latter is
to adapt the dual network to target appearance and shape variation.

\begin{figure}[t]
	\centering
	\includegraphics[width=0.49\textwidth,height=0.48\linewidth]{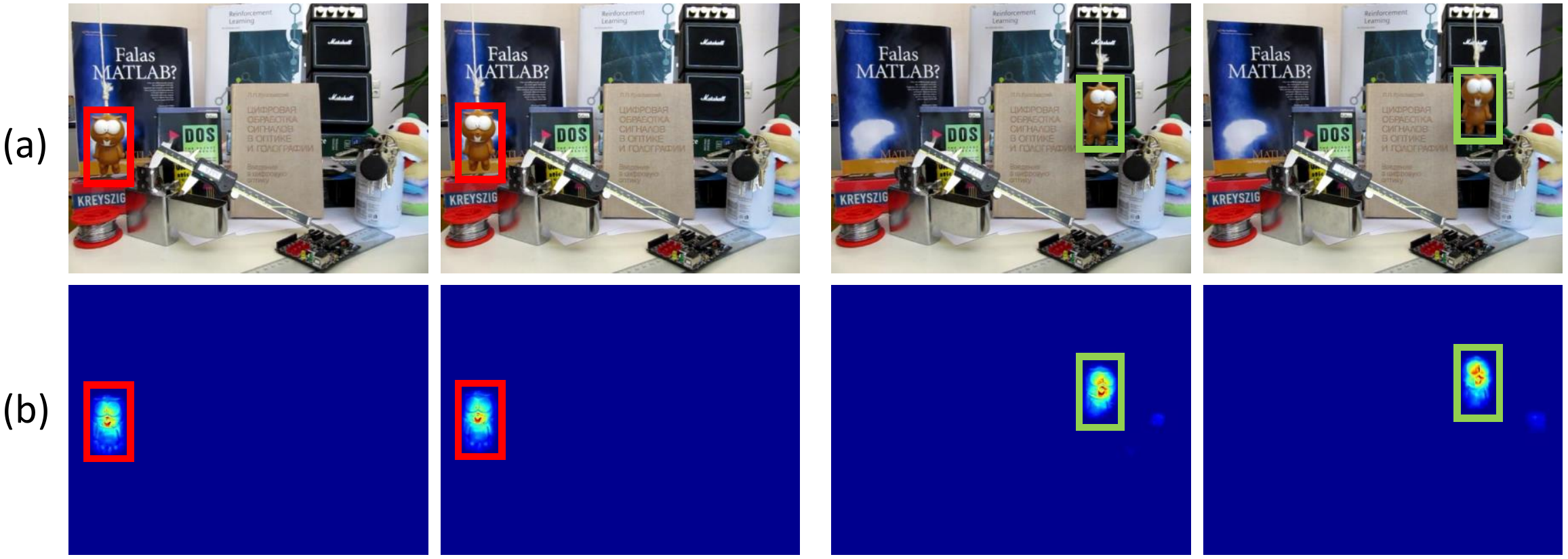}
	\vspace{-.5cm}
	\caption{Stochastic online update. (a) 1-st and 10-th, 200-th and 209-th frames in \texttt{Lemming}.
		(b) Thresholded target feature maps $\overline{\textbf{g}_D}(p_K)$ that are used in \eqref{loss_stochastic} for subsequent update.
	}\label{unsupervise}
	\vspace{-4mm}
\end{figure}

\smallskip

\noindent \textbf{Stochastic Update.}
Figure~\ref{unsupervise} shows some intermediate results during the online update of the stochastic stage.
When the target object is occluded by unknown objects or drifts to similar distracting objects, \textit{i.e.}, $\widehat{C}^{*}_t$ satisfies~\eqref{detect_occlusion},
we update the dual network using the best tracked patch $p_K$ in the last $K$ frames at time $t$, the current base patch $p_{t}$ as well as the random patches $\widetilde{p}^{(i)}_{t}$.
%
%
%
Note that by using the Gaussian target map $\textbf{t}(\cdot)$ as labels from different source patches, we train the dual network in a unique way.

Let $ \overline{\textbf{g}_D}(\cdot)$ denote the object appearance model
by thresholding the object of interest out of $\textbf{g}_D(\cdot)$.
Clearly the appearance model of the current target
$ \overline{\textbf{g}_D}(p_t)$
is similar to that of the last best tracked target $ \overline{\textbf{g}_D}(p_K)$. We refer this as "self-supervised learning" because except updating the dual net with Gaussian maps, we also update the last best tracked target.
The appearance model, or dual network, is refined by minimizing the following objective function:
%
\begin{gather}
  \hspace{-3.4cm}\ell(p_K, p_t,\mathit{\tilde{p}}^{(i)}_t) = ~\mathcal{D} \big( \overline{\textbf{g}_D}(p_t), \overline{\textbf{g}_D}(p_K) \big) \nonumber \\
  + (1-\phi_t) \mathcal{D} \big( \textbf{g}_D(p_t), \textbf{t}(p_t) \big)
  + \mathcal{D} \big( \textbf{g}_D(\tilde{p}^{(i)}_t), \textbf{t}(\tilde{p}^{(i)}_t) \big), \label{loss_stochastic}\\
\arg \min_\textbf{W}  ~\frac{1}{N}\sum_{i=1}^{N}\ell(p_K, p_t,\mathit{\tilde{p}}^{(i)}_t)+\beta{\|\mathbf{W}\|}_2^2,\label{update_num1}
\end{gather}
where $\mathcal{D}$ is the Euclidean distance function \eqref{similarity},
%
%
$\phi_{t}$ is a target mask indicating whether the $r$-th candidate $Z_t^{(r)}$
belongs to the target location: $\phi_{t}=1$~if~$s(Z_t^{(r)})\in{\text{target}}$, $\phi^t=0$ otherwise.
In addition, and $\beta$ and \textbf{W} are defined in \eqref{eq:loss_func}.
%

%

The first term in~\eqref{loss_stochastic} implies that we only choose the best tracked target from the last $K$ frames.
%
%
The second term indicates that the appearance model of background region is only updated in the current frame.
When the network is updated according to~\eqref{update_num1}, there must be occlusions or drifts near the target object.
As such, the estimated location is not updated since it is not reliable.
The third term corresponds to the loss of random patches which are center-shifted.
%
%
Overall,
the combination of different loss functions helps better separate targets from the background and suppress clutters or distractors near the objects of interest.

\medskip

\noindent \textbf{Periodic Update.}
In addition to the stochastic update, we refine the dual network at a fixed interval (e.g., 10 frames in this work) during the whole tracking process.
%
Note that we take the first frame into account because rich information in the first frame should be reserved in the learned model.
%
Therefore, the loss function is designed to constrain the network to approximate target appearance in a fully supervised manner using the Gaussian target map $\textbf{t}(\cdot)$ as labels.
The objective function is defined by
\begin{gather}
\ell(p_K, p_1) =
 \mathcal{D} \big( \textbf{g}_D(p_K), \textbf{t}(p_K) \big)
+ \mathcal{D} \big( \textbf{g}_D(p_1), \textbf{t}(p_1) \big),
\\
\arg \min_{\textbf{W}} \ell(p_K, p_1) +\beta{\|\mathbf{W}\|}_2^2. \label{update_num2}
\end{gather}
The dual network is updated periodically based on~\eqref{update_num2}
by taking the first frame and last best tracked frame into consideration in order to make the deep model more adept to target appearance.
\section{Experiments}\label{sec:experiment}

\begin{table*}[t]
	\scriptsize
	\renewcommand{\arraystretch}{1.4}
	\begin{center}
		\caption{Average \texttt{precision} scores on different attributes in the OPE experiment for OTB50. The best and  second  results are in \textcolor[rgb]{1,0,0}{\textbf{red}} and
			\textcolor[rgb]{0,1,0}{green}, respectively.
			The number in the parenthesis indicates the number of sequences involved in the corresponding attribute.
			}\label{precision}
		\vspace{0.1cm}
        \begin{tabular}{!{}c|c c c c c c c c c c c c!{}}
			\toprule
			&RTT &CNT &SMT &DSST &KCF &TGPR &MEEM &EBT &SCM &HCF &FCNT &DNT \\
			\midrule
			Illumination variation (25) &0.698 &0.566 &0.780 &0.741 &0.728 &0.671 &0.766 &0.814 &0.594 &\color{green}0.844 &0.830 &\color{red}0.876 \\
			Out-of-plane rotation (39) &0.767 &0.672 &0.830 &0.732 &0.729 &0.678 &0.840 &0.828 &0.618 &\color{green}0.869 &0.831 &\color{red}0.907 \\
			Scale variation (28) &0.721 &0.662 &0.827 &0.740 &0.679 &0.620 &0.785 &0.799 &0.672 &\color{green}0.880 &0.830 &\color{red}0.893\\
			Occlusion (29) &0.791 &0.662 &0.662 &0.725 &0.749 &0.675 &0.799 &0.574 &0.640 &\color{red}0.877 &0.797 &\color{green}0.850 \\
			Deformation (19) &0.814 &0.687 &0.858 &0.657 &0.740 &0.691 &0.846 &0.897 &0.586 &0.881 &\color{green}0.917 &\color{red}0.940 \\
			Motion blur (12) &0.687 &0.507 &0.745 &0.603 &0.650 &0.537 &0.715 &0.718 &0.339 &\color{red}0.844 &0.789 &\color{green}0.824 \\
			Fast motion (17) &0.719 &0.500 &0.723 &0.562 &0.602 &0.493 &0.742 &0.746 &0.333 &\color{green}0.790 &0.767 &\color{red}0.830 \\
			In-plane rotation (31) &0.713 &0.661 &0.836 &0.780 &0.725 &0.675 &0.800 &0.782 &0.597 &\color{green}0.868 &0.811 &\color{red}0.893\\
			Out-of-view (6) &\color{red}0.854 &0.502 &0.687 &0.533 &0.650 &0.505 &0.727 &0.0.729 &0.429 &0.695 &0.741 &\color{green}0.810\\
			Background cluttered(21) &0.770 &0.646 &0.789 &0.691 &0.753 &0.717 &0.797 &0.825 &0.578 &\color{red}0.885 &0.799 &\color{green}0.884\\
			Low resolution (4) &0.660 &0.557 &0.705 &0.534 &0.381 &0.538 &0.490 &0.713 &0.305 &\color{red}0.897 &\color{green}0.765 &0.719\\
			\midrule
			Average &0.660 &0.723 &0.852 &0.739 &0.740 &0.705 &0.830 &0.846 &0.649 &\color{green}0.891 &0.856 &\color{red}0.907\\
			\bottomrule
		\end{tabular}
	\end{center}
	\vspace{-1mm}
\end{table*}

%
\begin{table*}[t]
	\scriptsize
	\renewcommand{\arraystretch}{1.4}
	\centering
	\caption{Average \texttt{success} scores on different attributes in the OPE experiment for OTB50. The best and second results are in \textcolor[rgb]{1,0,0}{\textbf{red}} and
		\textcolor[rgb]{0,1,0}{green}, respectively.
		The number in the parenthesis indicates the number of sequences involved in the corresponding attribute.
		}\label{success}
	\vspace{-.2cm}
	\begin{tabular}{!{}c|c c c c c c c c c c c c!{}}
		\toprule
		&RTT &CNT &SMT &DSST &KCF &TGPR &MEEM &EBT &SCM &HCF &FCNT &DNT \\
		\midrule
		Illumination variation (25) &0.539 &0.456 &0.556 &0.506 &0.493 &0.484 &0.533 &0.563 &0.473 &0.560 &\color{green}0.598 &\color{red}0.650 \\
		Out-of-plane rotation (39) &0.566 &0.501 &0.582 &0.491 &0.495 &0.485 &0.558 &0.572 &0.470 &\color{green}0.587 &0.581 &\color{red}0.661 \\
		Scale variation (28) &0.527 &0.508 &0.513 &0.451 &0.427 &0.418 &0.498 &0.533 &0.518 &0.531 &\color{green}0.558 &\color{red}0.642 \\
		Occlusion (29) &0.585 &0.503 &0.563 &0.480 &0.514 &0.484 &0.552 &0.568 &0.487 &\color{green}0.606 &0.571 &\color{red}0.641 \\
		Deformation (19) &0.628 &0.524 &0.640 &0.474 &0.534 &0.510 &0.560 &0.609 &0.448 &0.626 &\color{green}0.644 &\color{red}0.690 \\
		Motion blur (12) &0.541 &0.417 &0.565 &0.458 &0.497 &0.434 &0.541 &0.544 &0.298 &\color{green}0.616 &0.580 &\color{red}0.634 \\
		Fast motion (17) &0.549 &0.404 &0.545 &0.433 &0.459 &0.396 &0.553 &0.561 &0.296 &\color{green}0.578 &0.565 &\color{red}0.635 \\
		In-plane rotation (31) &0.529 &0.495 &0.571 &0.532 &0.497 &0.479 &0.535 &0.539 &0.458 &\color{green}0.582 &0.555 &\color{red}0.645\\
		Out-of-view (6) &\color{red}0.672 &0.439 &0.571 &0.490 &0.550 &0.442 &0.606 &0.562 &0.361 &0.575 &0.592 &\color{green}0.636 \\
		Background cluttered (21) &0.598 &0.488 &0.593 &0.492 &0.535 &0.522 &0.569 &0.574 &0.450 &\color{green}0.623 &0.564 &\color{red}0.652\\
		Low resolution (4) &0.508 &0.437 &0.461 &0.352 &0.312 &0.370 &0.360 &0.472 &0.279 &\color{red}0.557 &0.514 &\color{green}0.536\\
		\midrule
		Average &0.588 &0.545 &0.597 &0.505 &0.514 &0.503 &0.566 &0.586 &0.499 &\color{green}0.605 &0.599 &\color{red}0.664\\
		\bottomrule
	\end{tabular}
	\vspace{-1mm}
\end{table*}

In this section, we discuss the implementation details and experimental settings of the proposed DNT tracker.
%
Quantitative and qualitative results on two large-scale tracking benchmark data sets (OTB50~\cite{wu2013online}, OTB100~\cite{wu2015object} and VOT2015~\cite{kristan2015visual}) are presented.
%

\subsection{Implementation Details}
%
\begin{figure}[t]
\begin{center}
\begin{tabular}{c@{}c}
\vspace{1mm}
\hspace{-5mm}
\includegraphics[width=0.5\linewidth]{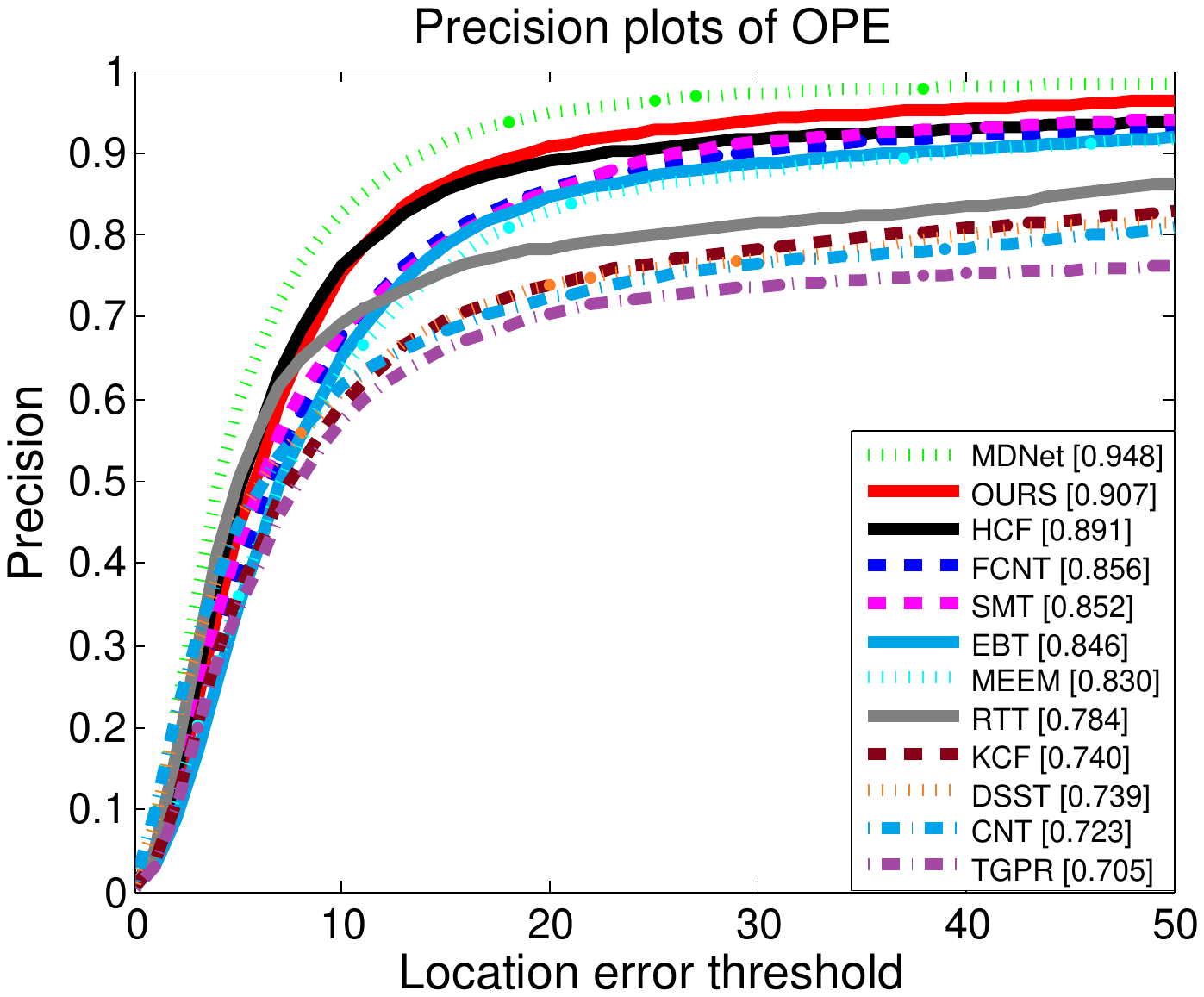}
\includegraphics[width=0.5\linewidth]{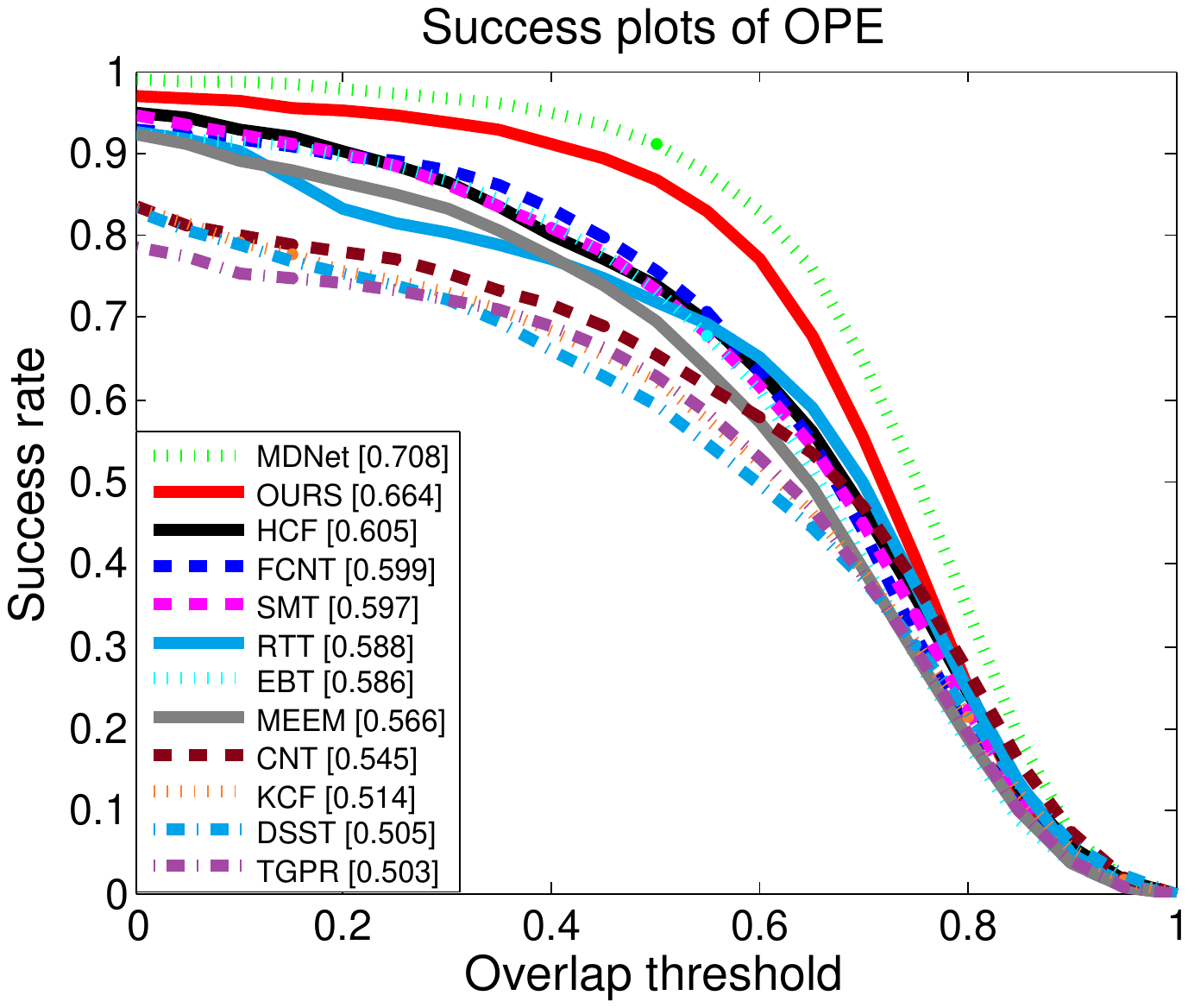} \\
\vspace{1mm}
\hspace{-5mm}
\includegraphics[width=0.5\linewidth]{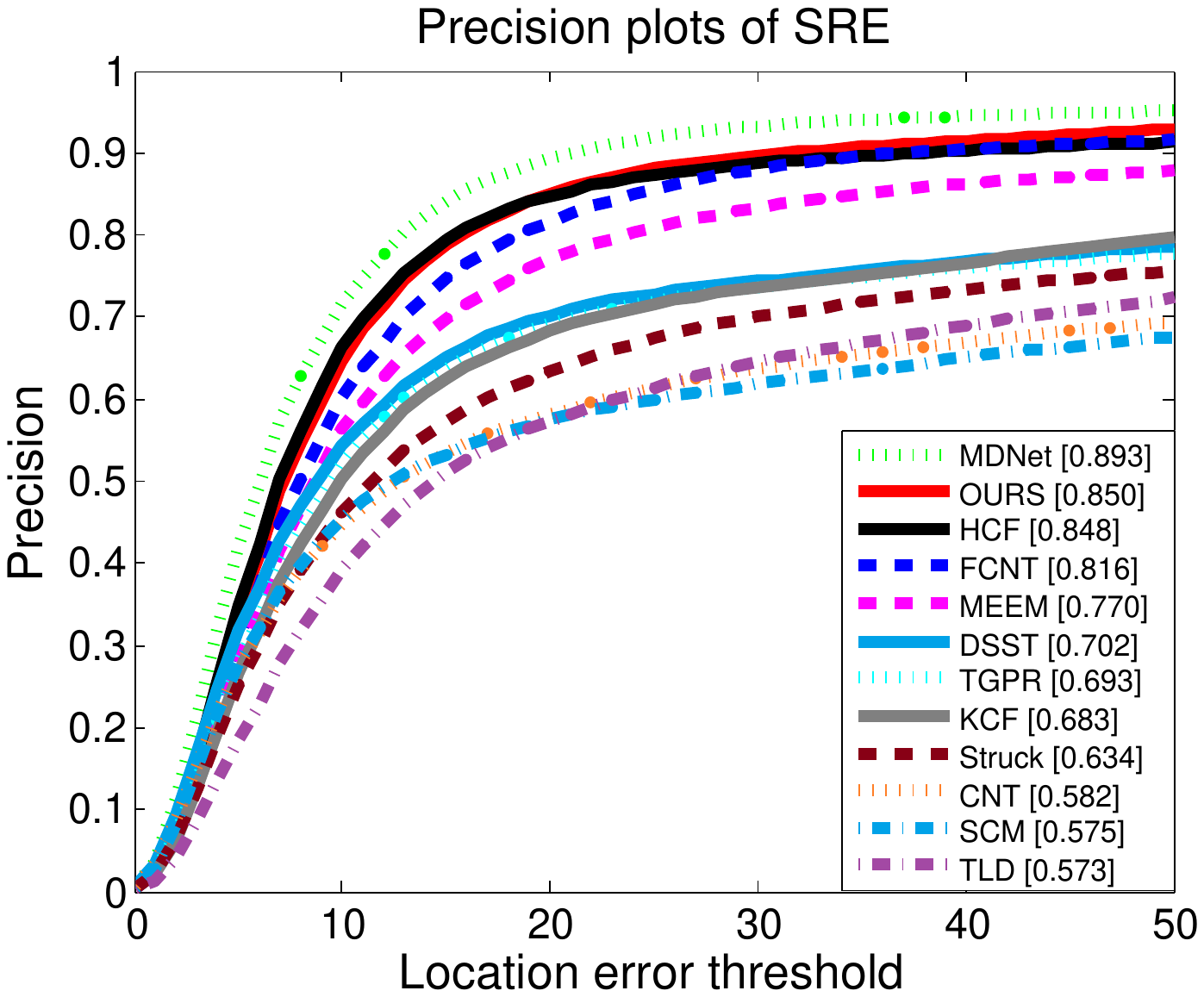}
\includegraphics[width=0.5\linewidth]{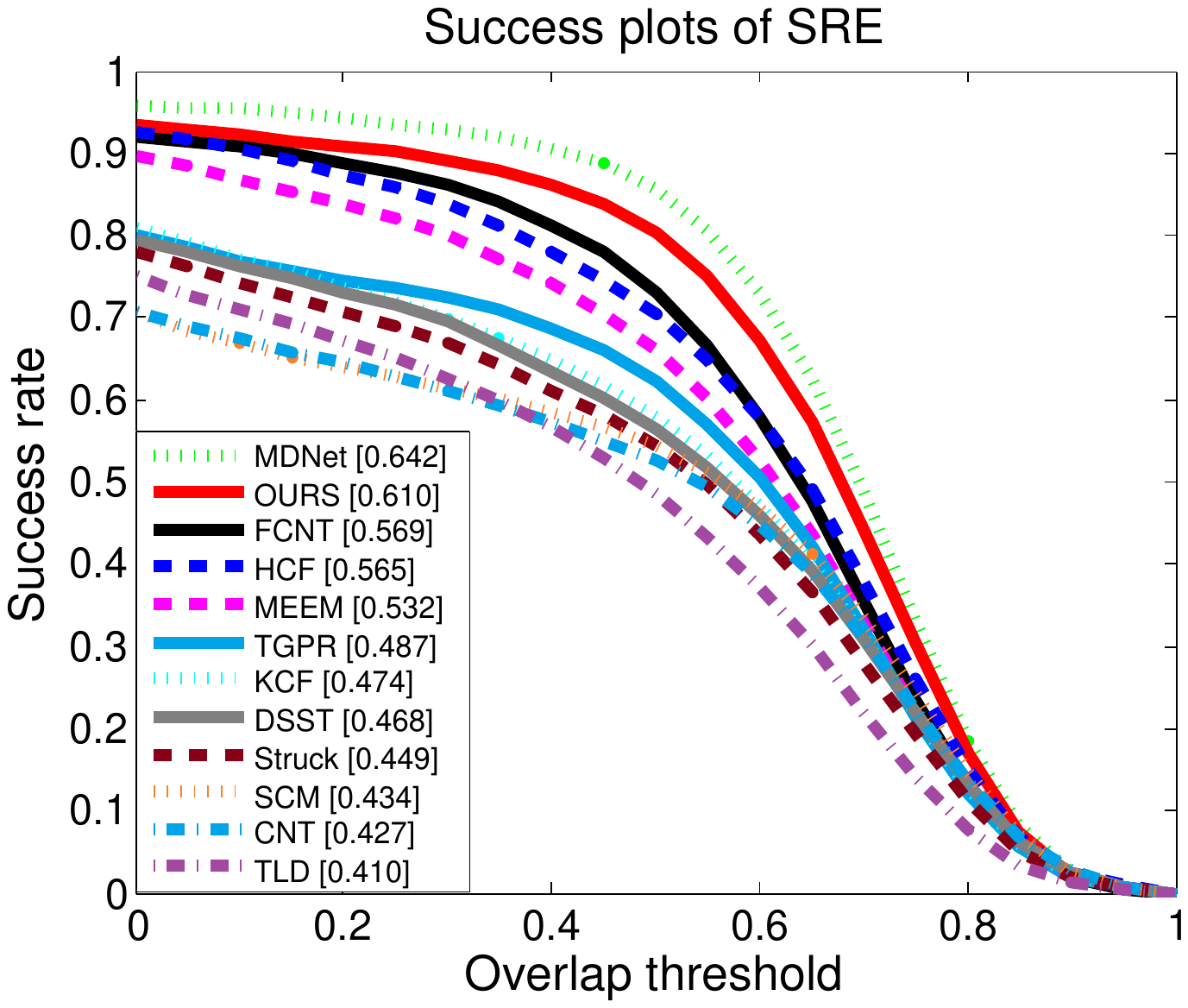} \\
\hspace{-4mm}
\includegraphics[width=0.5\linewidth]{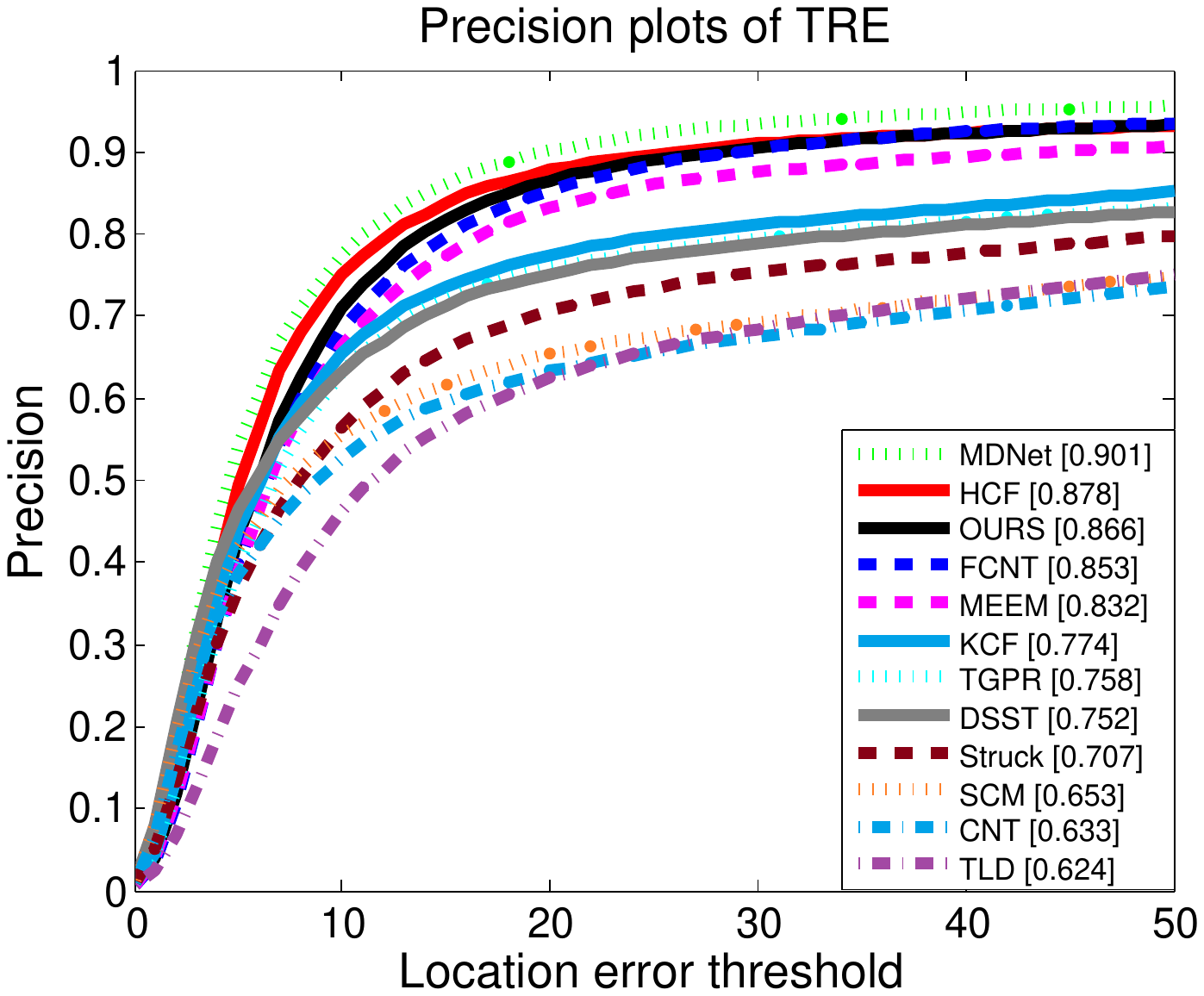}
\includegraphics[width=0.5\linewidth]{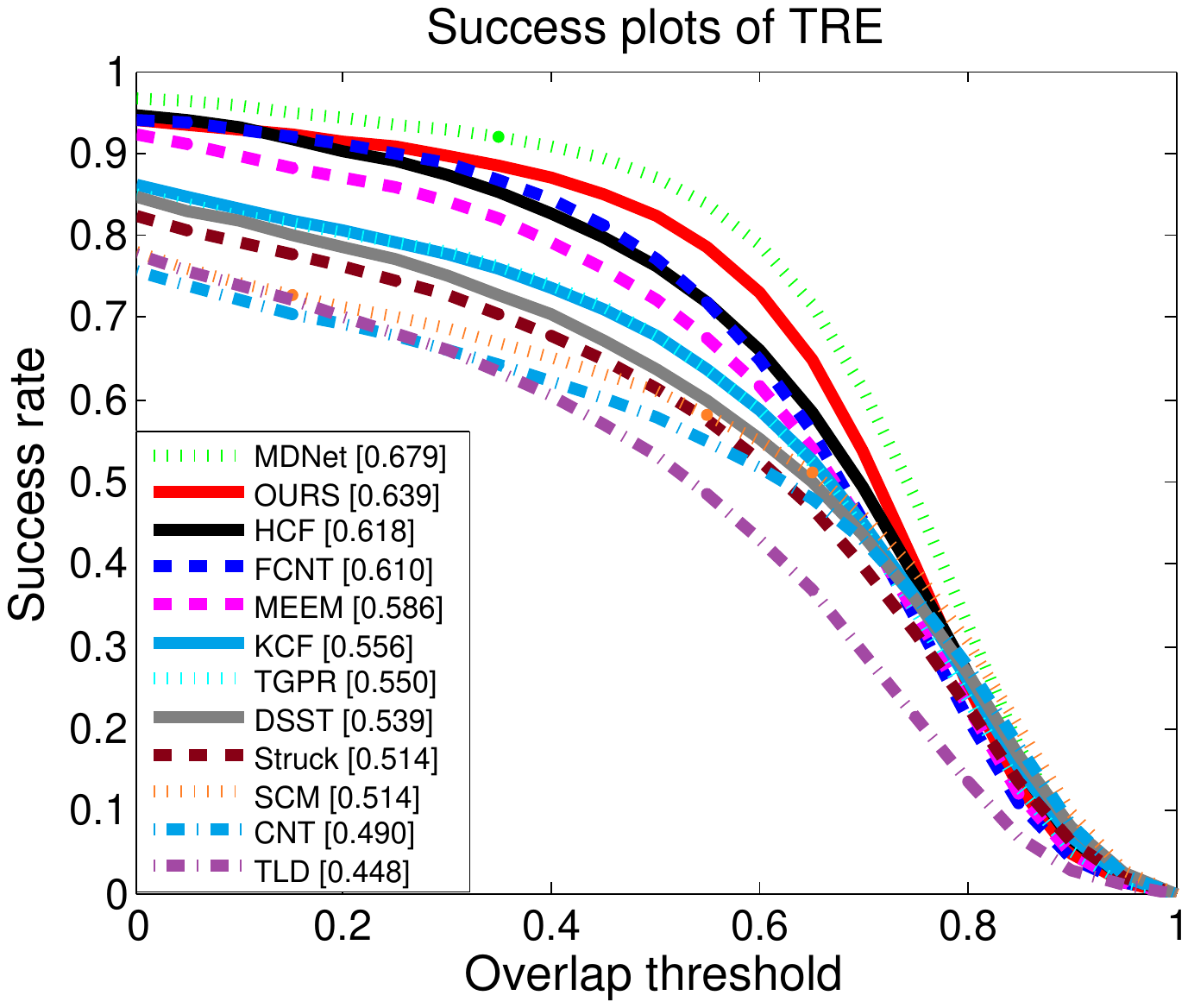}
\end{tabular}
\vspace{-4.5mm}
\end{center}
\caption{Precision (left) and overlap (right) plots for the 11 trackers over the 50 benchmark sequences in OTB50 using one-pass evaluation (OPE), spatial robustness evaluation (SRE) and temporal robustness evaluation (TRE). The
performance score of precision plot is at error threshold of 20 pixels while the performance score of success plot is the AUC value. Our DNT tracker performs favorably against state-of-the-arts.}
\label{fig:precision}
\vspace{-3mm}
\end{figure}

Our DNT tracker is implemented based on Caffe \cite{jia14} and runs at 5 frames per second
on a PC with a 3.4GHz CPU and a TITAN GPU.
The dual network is trained online with SGD for target localization using the first frame for 50 iterations. The initial learning rate is $10^{-6}$ with a momentum of 0.6.
To alleviate overfitting, the weight decay~$\beta$ in the objective functions is set uniformly to be 0.005.
At each frame, 600 target candidates are randomly sampled.
The variance of candidate location parameters are set to \{10,~10,~0.01\} for translation and scale, respectively. We observe that the final target localization is not sensitive to $\lambda$ in~\eqref{maxconfidence}. In early experiments, $\lambda$ is set from 0.2 to 0.8 and the performance alters within 2\% in OTB50~\cite{wu2013online}. Thus we set it 0.4 to pay a little more attention to semantic information in higher layer.
$\theta$ in~\eqref{detect_occlusion} is set 0.45 empirically through experiments to handle occlusion and drifts. In each stochastic update, $K=10$ is chosen as the best tracked patch $p_K$ in the last $K$ frames.
We fix all the parameters fixed throughout the experiments and data sets.

\subsection{Evaluation on OTB Data Set}

{\flushleft{\textbf{Data Set and Evaluation Settings.}}}
The OTB50 data set~\cite{wu2013online} includes 50 sequences with 11 various challenging factors such as illumination variation, deformation, motion blur, scale variation, etc.
We compare with the state-of-the-art trackers, including Struck~\cite{hare2011}, SCM~\cite{zhong14}, TLD~\cite{kalal12},
MDNet~\cite{nam2015learning}, EBT~\cite{zhu2016robust}, CNT~\cite{zhang2016robust}, DSST~\cite{danelljan2014accurate}, KCF~\cite{henriques15}, TPGR~\cite{gao2014transfer}, MEEM~\cite{zhang2014meem}, RTT~\cite{cui2016recurrently}, FCNT~\cite{wangvisual}, SMT~\cite{hong2015tracking} and HCF~\cite{machao15},
based on the precision and success scores.
One-pass evaluation (OPE), spatial robustness evaluation (SRE) and temporal robustness evaluation (TRE) are all conducted to thoroughly evaluate our tracker. Among them, SRE randomizes the initial bounding box by perturbation and TRE randomizes the the starting frame of the sequence. As additional evaluations to OPE, SRE and TRE can better demonstrate trackers' robustness.

All the trackers are ranked according to the center location error threshold of 20 pixels and the area under curve (AUC) of success rate plot.
The precision plot demonstrates the percentage of frames where the distance between the estimated target location and ground truth location is within a given threshold. Whereas the success plot illustrates the percentage of frames where the overlap ratio between the estimated bounding box and ground truth bounding box is higher than a threshold $\tau \in [0, 1]$.
\begin{table}[t]
	\centering
	\caption{Time complexity(frames per second) of different DCNN based trackers. Note that HCFT is fast due to a filter-based tracker scheme under FFT domain. }\label{timecomplexity}
	\footnotesize{
		\begin{tabular}{c@{}c}
\hspace{-3mm}
        \includegraphics[width=\linewidth,height=0.08\linewidth]{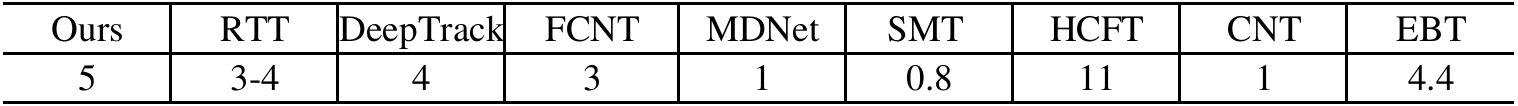} \\
		\end{tabular}
	}
\vspace{-0.5cm}
\end{table}

{\flushleft{\textbf{Quantitative Evaluation.}}}
Figure~\ref{fig:precision} demonstrates the average precision plots and success plots on all the 50 sequences of the top eleven trackers. Our tracker performs favorably among the state-of-the-arts, which achieves 1.8\% and 9.75\% gain over the state-of-the-art trackers on precision and overlap scores in the OPE.
Among the three experiments, SRE is the most difficult since target object is initialized with inaccurate locations. Though all the trackers achieve worse performance, our tracker can still estimate location and scale accurately with salient semantics. In the TRE experiment, dual net is initially trained in different frames, our tracker performs worse than HCF~\cite{machao15} because target object is missing or partly occluded in some intermediate frames. The dual net cannot capture the semantic information of target object. From the figures, we observe that the success plots of our method in all the three experiments are the highest, which demonstrates the effectiveness of the scale estimation in the DNT tracker. As is known to all, the success rate plot is much more convincing than the precision plot in the OTB data set because both location and scale are evaluated.

Table~\ref{timecomplexity} shows the time complexity of different Deep ConvNet based trackers. In our method, most time are spent in random patch update when occlusion or drift occurs. There can be a significant increase in speed without random patch update, but the precision will be decreased accordingly. HCFT~\cite{machao15} and some other correlation filter based trackers are fast because they are calculated in the Fourier domain with Fast Fourier Transform (FFT).  They can be regarded as an extension of KCF~\cite{henriques15} tracker.

We also evaluate the tracker performance in terms of individual attributes in Table \ref{precision} and \ref{success}.
%
%
In terms of precision score, Table~\ref{precision} shows that the proposed DNT
algorithm performs favorably against the state-of-the-art methods
in most cases, such as scale variation, rotation, fast motion, etc.
However, the DNT method performs slightly worse than other trackers
in occlusion (\emph{Subway}, \emph{Walking2}, \emph{Freeman4}), motion blur (\emph{Jumping}, \emph{Woman}) and background cluttered (\emph{Soccer}, \emph{Matrix}, \emph{Ironman}).
The main reason is that activation values of target object are equal to or even lower than the surrounding background.

In terms of success score, Table~\ref{success} shows the DNT algorithm performs well against the state-of-the-art methods except when the image resolution is low
(\emph{Ironman}, \emph{CarDark}, \emph{Walking2}).
In these sequences, boundary information
creates a weak prior map as reference for ICA-R, which leads to an inaccurate localization in tracking.
%

\begin{figure}[t]
\begin{center}
\begin{tabular}{c@{}c}
\vspace{1mm}
\hspace{-4mm}
\includegraphics[width=0.5\linewidth]{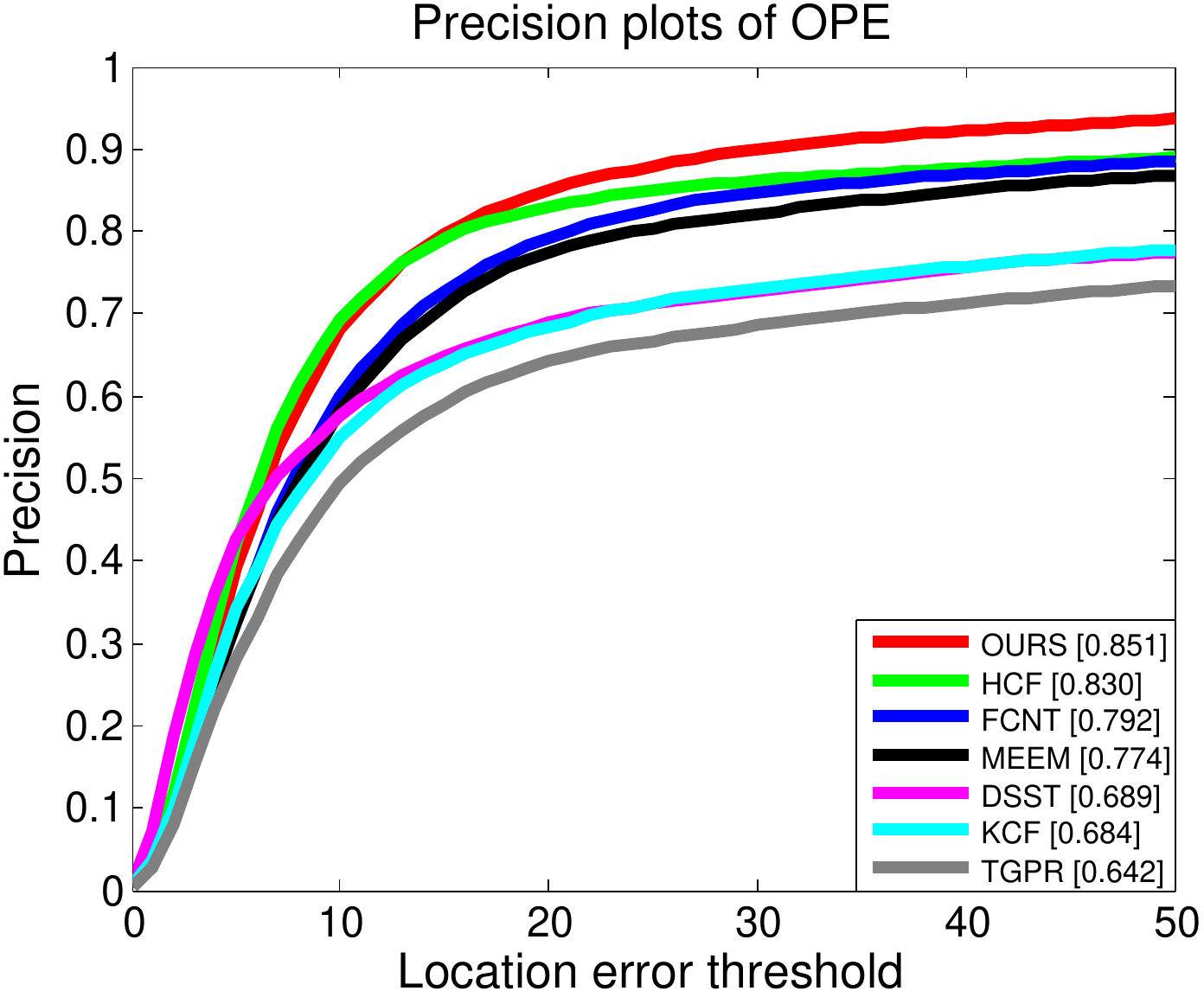}
\includegraphics[width=0.5\linewidth]{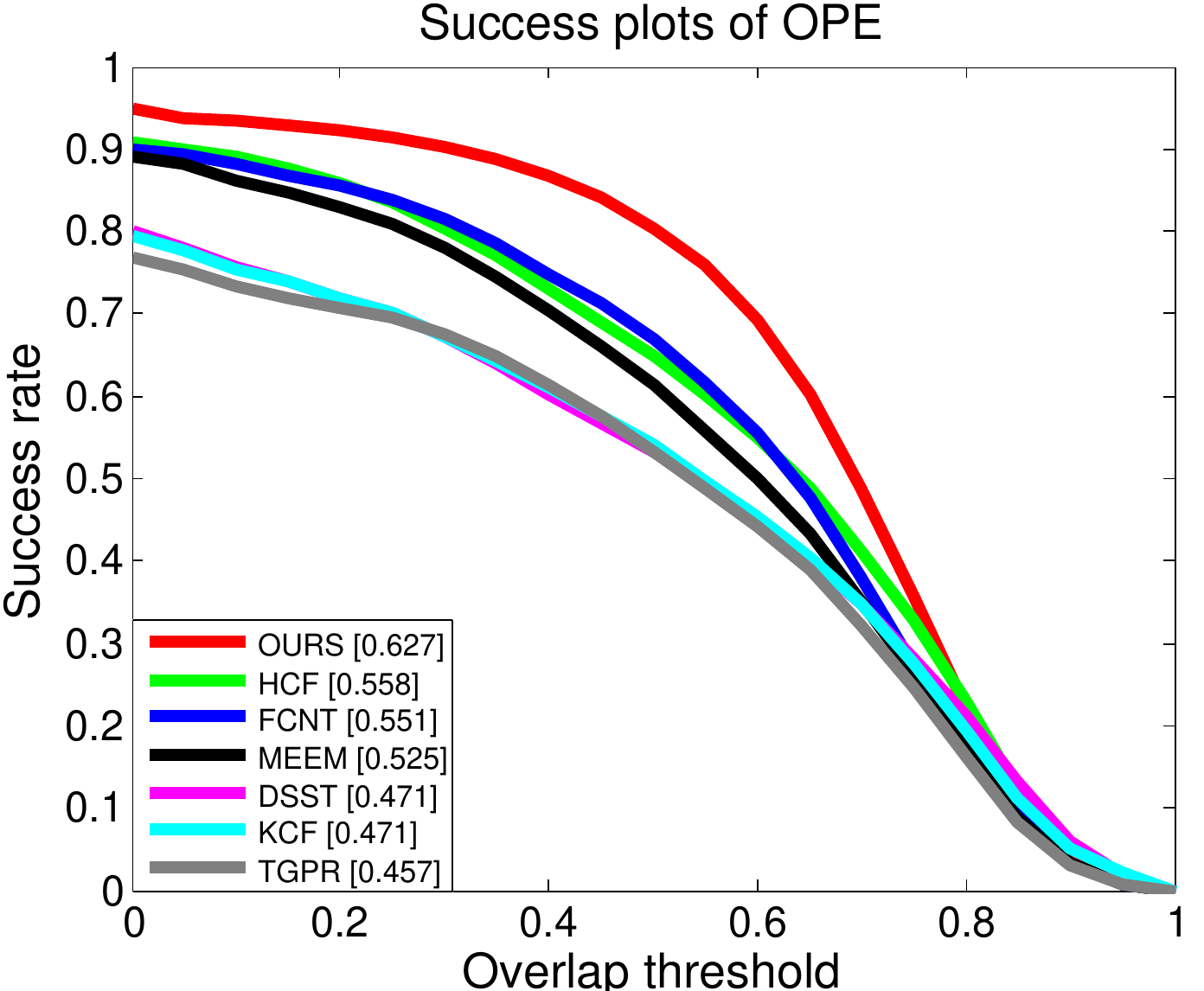} \\
\vspace{1mm}
\hspace{-4mm}
\includegraphics[width=0.5\linewidth]{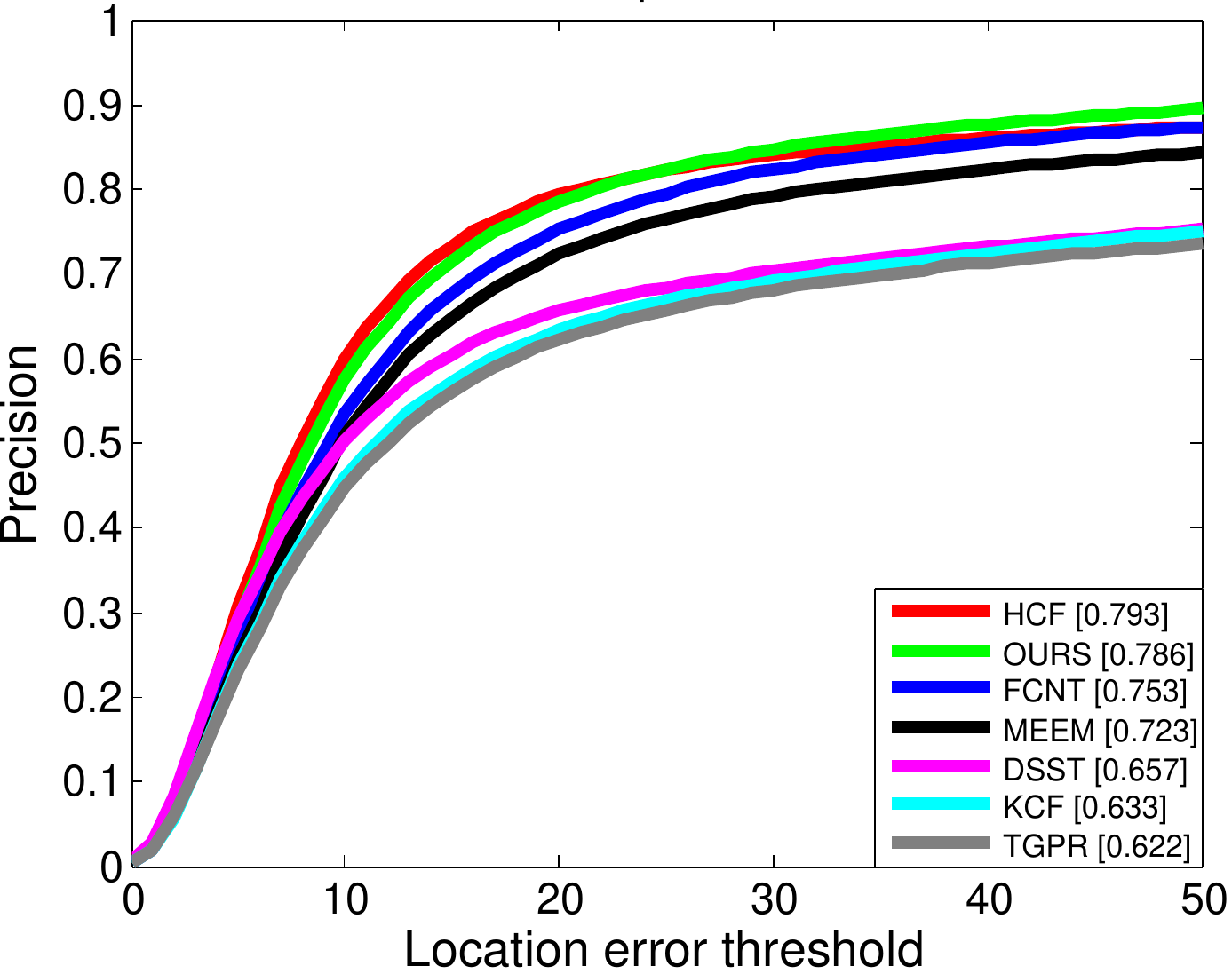}
\includegraphics[width=0.5\linewidth]{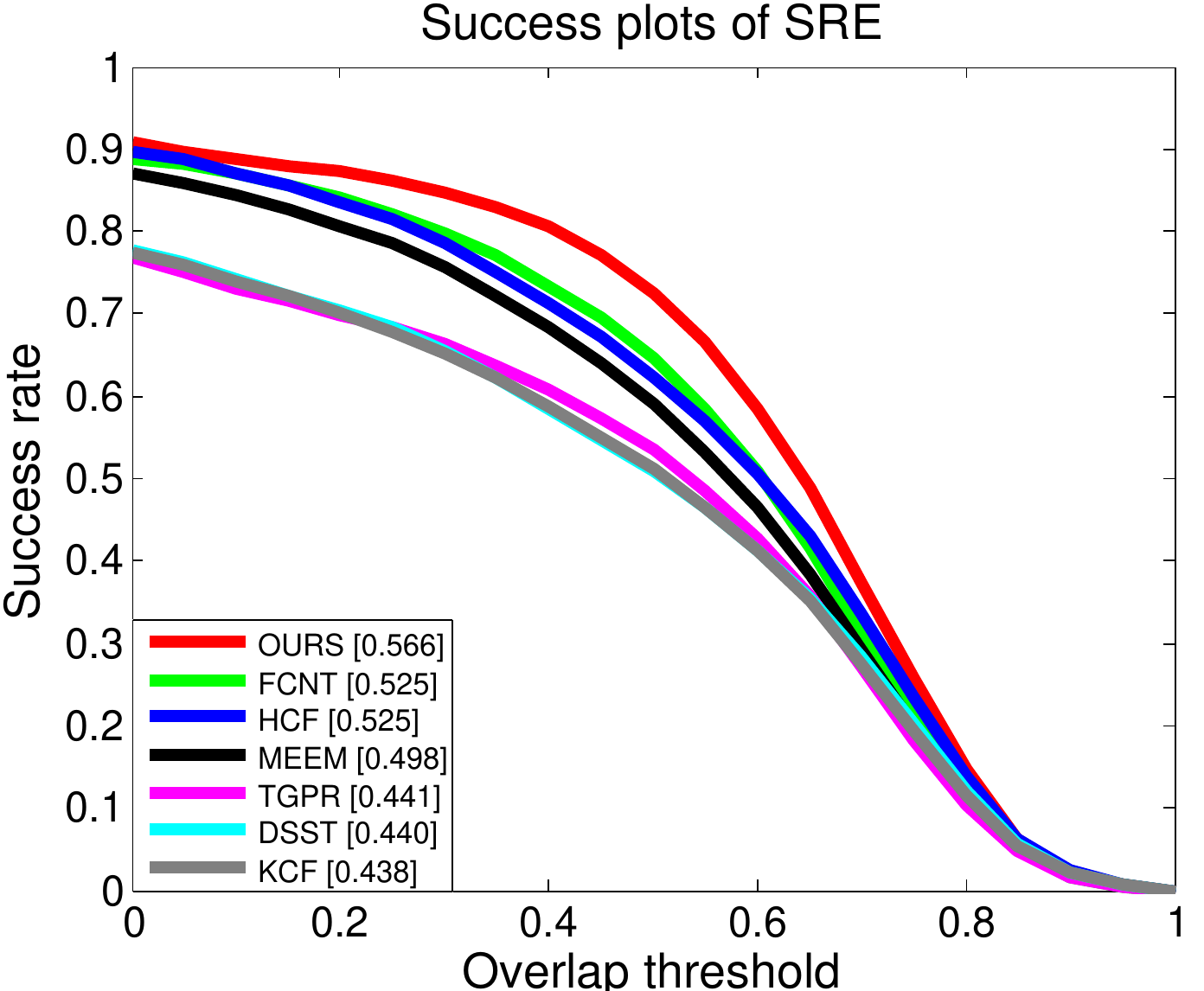} \\
\hspace{-4mm}
\includegraphics[width=0.5\linewidth]{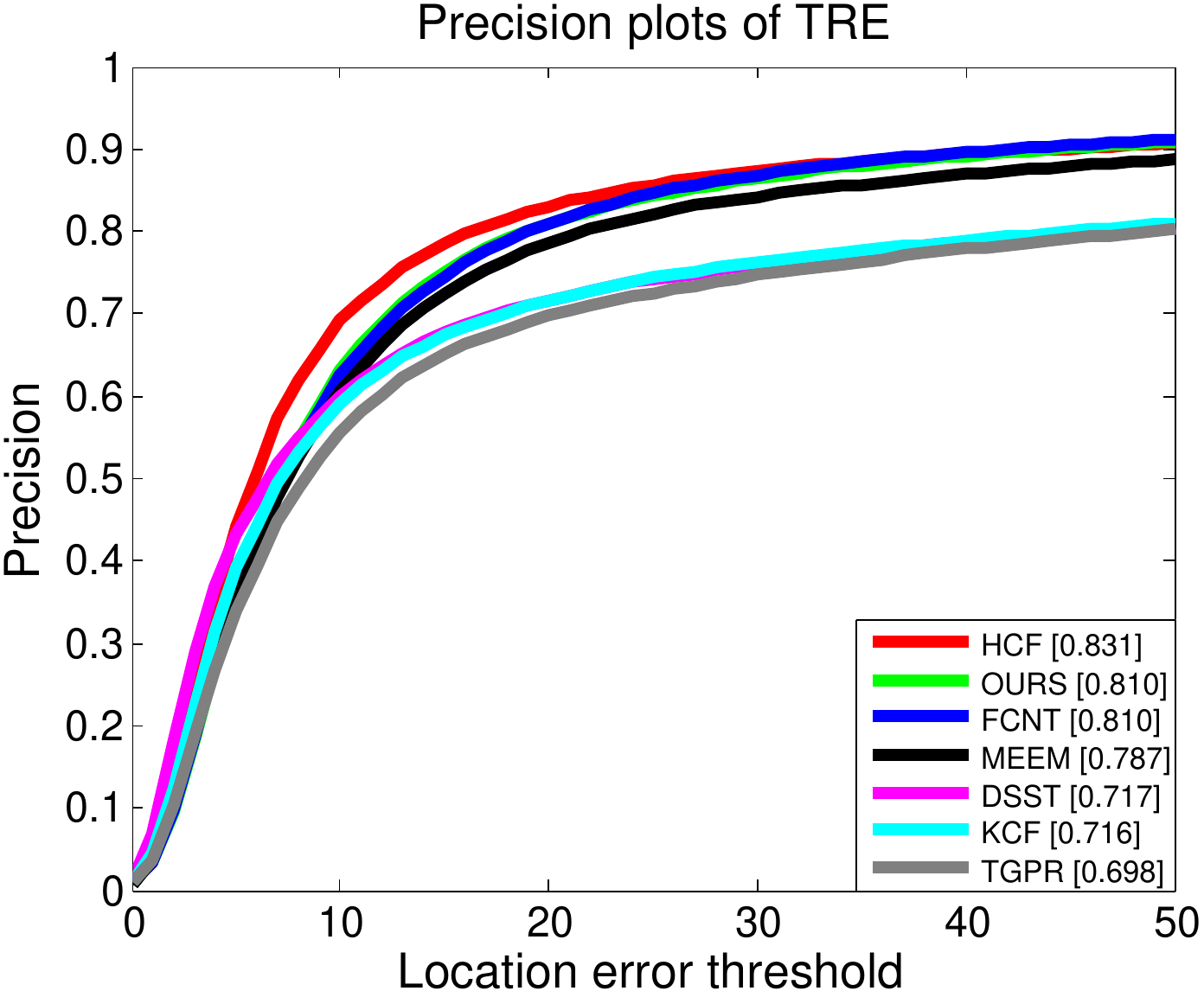}
\includegraphics[width=0.5\linewidth]{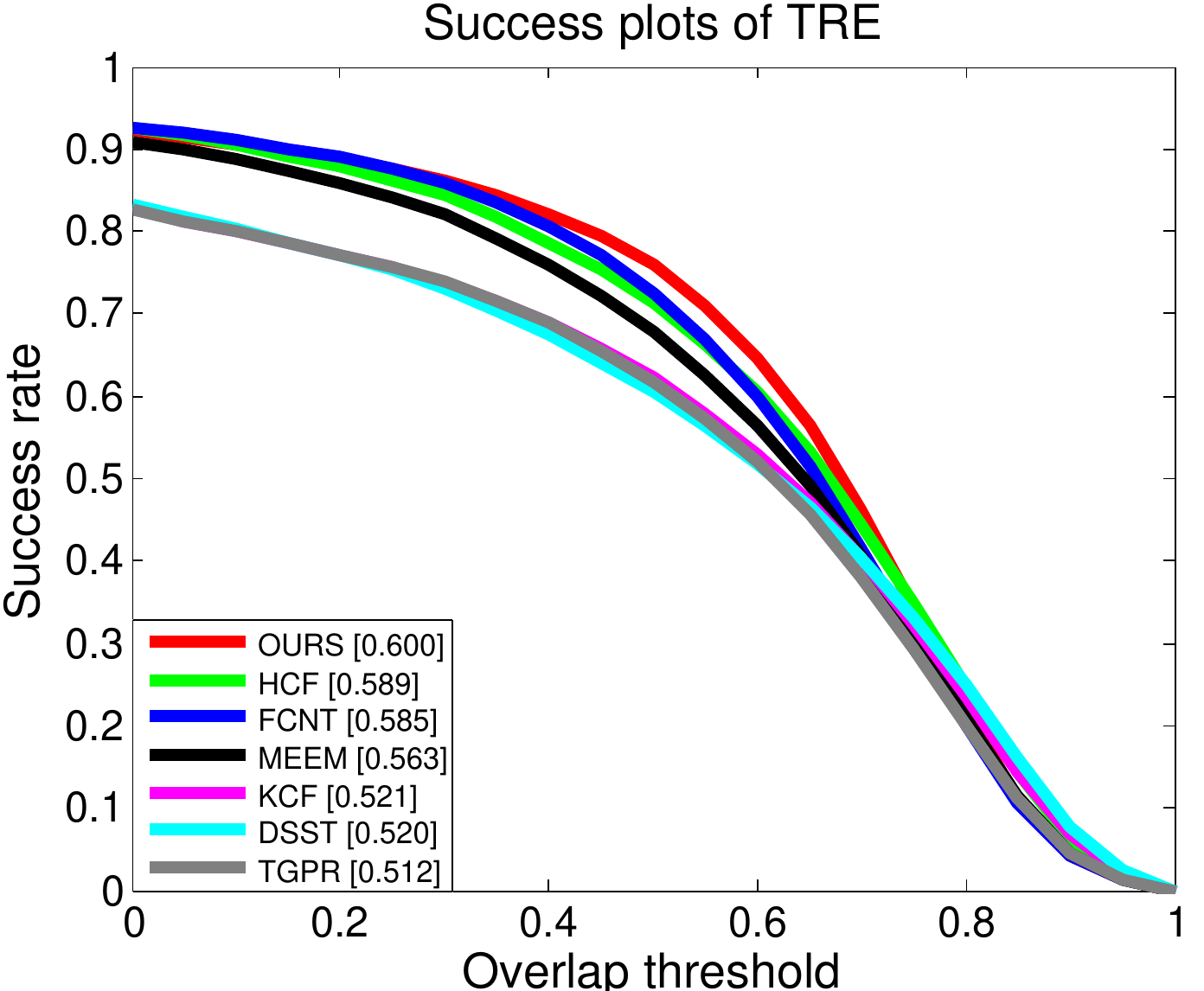}
\end{tabular}
\vspace{-4mm}
\end{center}
\caption{Precision (left) and overlap (right) plots for the 7 trackers over the 100 sequences in OTB100 using one-pass evaluation (OPE), spatial robustness evaluation (SRE) and temporal robustness evaluation (TRE). The performance score of precision plot is at error threshold of 20 pixels while the performance score of success plot is the AUC value.}
\label{fig:precision100}
\vspace{-3mm}
\end{figure}

\begin{figure}[h]
\begin{center}
\begin{tabular}{c@{}c}
\hspace{-3mm}
\includegraphics[width=1\linewidth, height=0.65\linewidth]{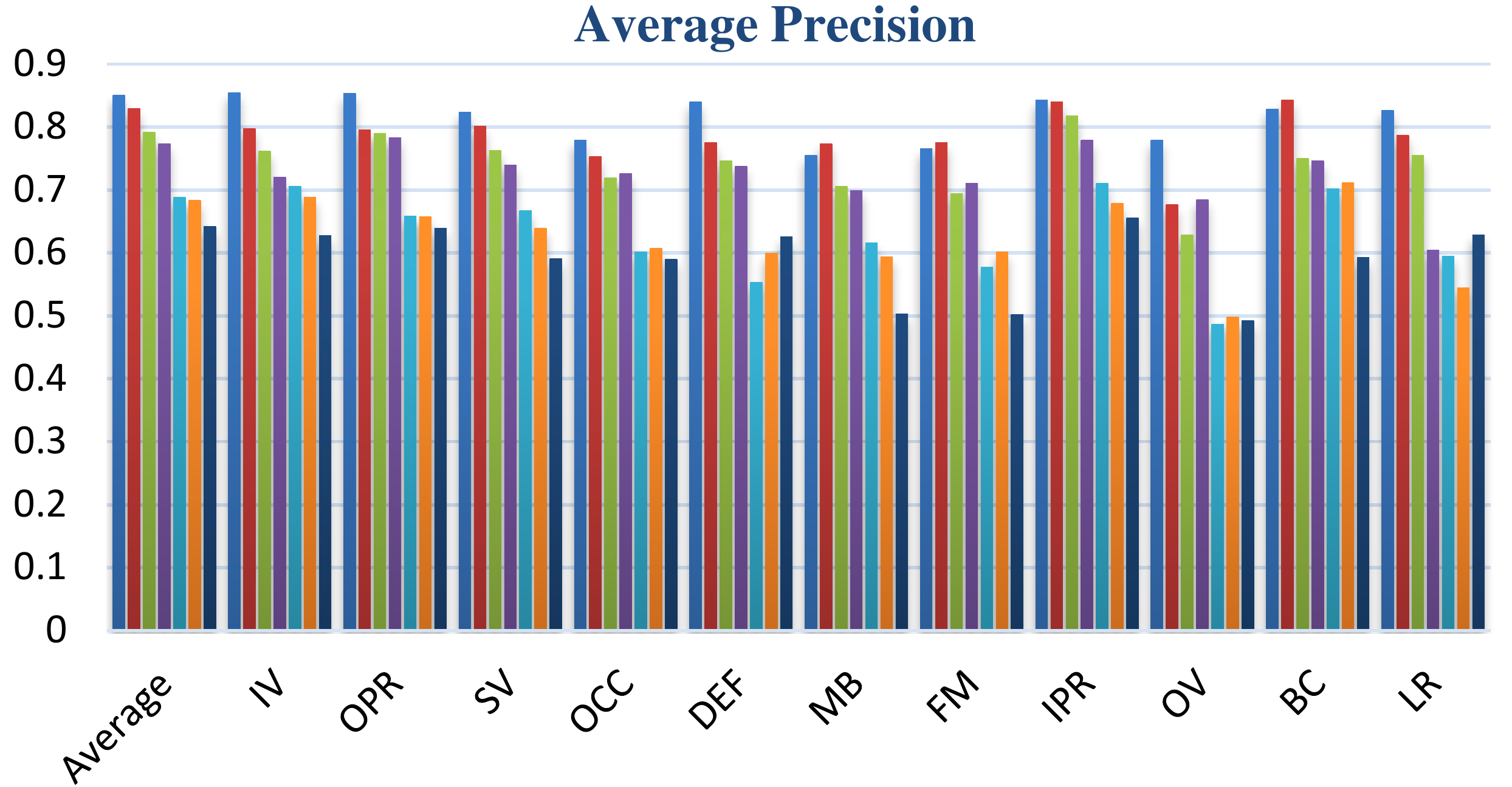}\\
\hspace{-3mm}
\includegraphics[width=1\linewidth, height=0.65\linewidth]{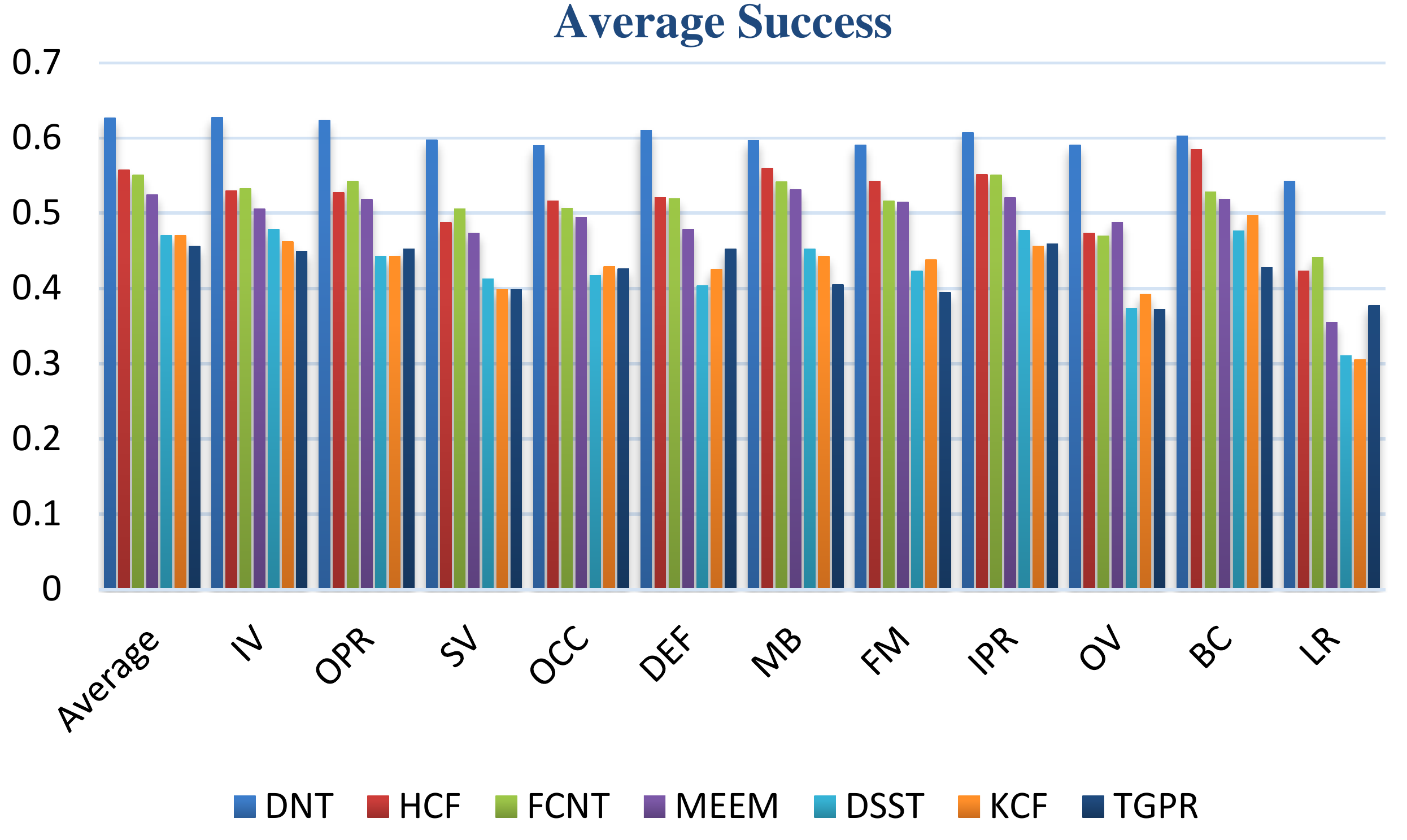} \\
\end{tabular}
\vspace{-4mm}
\end{center}
\caption{Average precision and success scores of the 7 trackers under different attributes of the sequences in OTB100 using one-pass evaluation (OPE), including illumination variation (IV), out-of-plane rotation (OPR), scale variation (SV), occlusion (OCC), deformation (DEF), motion blur (MB), fast motion (FM), in-plane rotation (IPR), out-of-view (OV), background cluttered (BC) and low resolution (LR).}
\label{fig:attribute}
\vspace{-3mm}
\end{figure}

\begin{table*}[h]
	\scriptsize
	\renewcommand{\arraystretch}{1.4}
	\centering
	\caption{The average ranks of accuracy and robustness under baseline and Overall experiments in VOT2015~\cite{kristan2015visual}. We rank all the trackers according to the VOT article report from top to down. The first, second and third best methods are highlighted in \first{red}, \second{blue}, \third{green} colors, respectively.}
	\label{vot2015}
\renewcommand{\multirowsetup}{\centering}
\begin{tabular}{>{\centering}m{1.5cm}|>{\centering}m{1.5cm} >{\centering}m{1.5cm} >{\centering}m{1.5cm}|>{\centering}m{1.5cm} p{1.5cm}<{\centering}}
\toprule
\multirow{2}{*}{\textbf{Trackers}} & \multicolumn{3}{ c }{\textbf{baseline}} \vline & \multicolumn{2}{ c }{\textbf{Overall}} \\\cline{2-6}
 & \textbf{Acc. Rank} & \textbf{Rob. Rank} & \textbf{Expected overlap} & \textbf{Acc. Rank} & \textbf{Rob. Rank} \\\hline
\textbf{DeepSRDCF} & \first{2.73} & \second{4.23} & \first{0.3057} & \first{2.73} & \second{4.23} \\
\textbf{EBT} & 7.35 & \first{3.80} & \second{0.3003} & 7.35 & \first{3.80} \\
\textbf{DNT} & \second{3.42} & \third{6.03} & \third{0.2748} & \second{3.42} & \third{6.03} \\
\textbf{LDP} & 5.68 & 6.50 & 0.2693 & 5.68 & 6.50 \\
\textbf{sPST} & \third{4.52} & 6.55 & 0.2570 & \third{4.52} & 6.55 \\
\textbf{struck} & 6.07 & 7.07 & 0.2387 & 6.07 & 7.07 \\
\textbf{s3tracker} & 4.25 & 8.08 & 0.2335 & 4.25 & 8.08 \\
\textbf{sumshift} & 5.08 & 8.12 & 0.2280 & 5.08 & 8.12 \\
\textbf{DAT} & 7.12 & 10.18 & 0.2128 & 7.12 & 10.18 \\
\textbf{MEEM} & 5.00 & 8.72 & 0.2106 & 5.00 & 8.72 \\
\textbf{RobStruck} & 5.78 & 7.22 & 0.2079 & 5.78 & 7.22 \\
\textbf{MCT} & 6.33 & 9.90 & 0.2076 & 6.33 & 9.90 \\
\bottomrule
\end{tabular}
\end{table*}

\begin{figure}[h]
\begin{center}
\begin{tabular}{c@{}c}
\hspace{-16mm}
\includegraphics[width=0.75\linewidth, height = 0.7\linewidth]{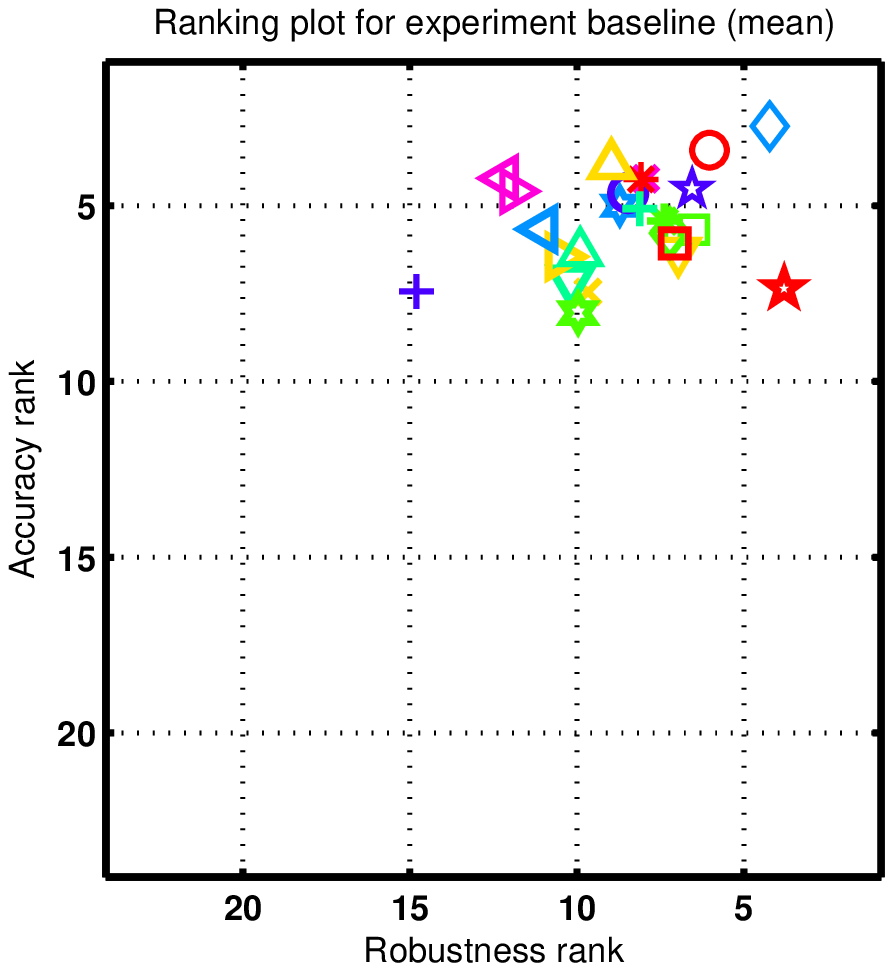} \\
\hspace{-16mm}
\includegraphics[scale=0.5]{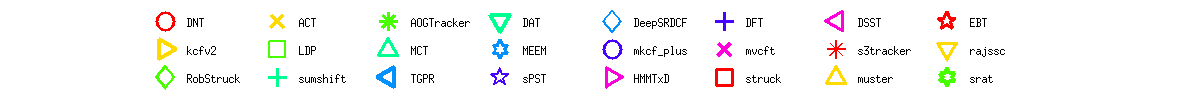}
\end{tabular}
\vspace{-4mm}
\end{center}
\caption{The robustness-accuracy ranking plots of the 24 leading trackers under the baseline evaluation in the VOT2015 data set. The better trackers are located at the upper-right corner.}
\label{vot_fig}
\vspace{-3mm}
\end{figure}

To gain more insights on the effectiveness of the proposed algorithm, we further report the performance of the top seven trackers (precision score > 0.7 and success score > 0.5 for the OPE experiment in OTB50) in the OTB100 data set~\cite{wu2015object} with 100 sequences (a supplementary of OTB50), including  DSST~\cite{danelljan2014accurate}, KCF~\cite{henriques15}, TPGR~\cite{gao2014transfer}, MEEM~\cite{zhang2014meem}, FCNT~\cite{wangvisual}, and HCF~\cite{machao15} and the proposed DNT tracker.
Figure~\ref{fig:precision100} demonstrates the average precision plots and success plots on the 100 sequences of the top seven trackers.
The OPE, SRE and TRE plots share the same spirit with the evaluations in OTB50. Our method can estimate both location and scale accurately compared with other trackers.

To facilitate more detailed analysis, we analyse the performance of these trackers on different attributes in figure~\ref{fig:attribute}. The DNT tracker can well handle various challenging factors and consistently outperform the other six trackers in almost all the attributes.

\begin{figure*}[t]
	\centering
	\vspace{2mm}
	\begin{tabular}{c}
\hspace{1.5mm}
		\includegraphics[width=0.6\linewidth]{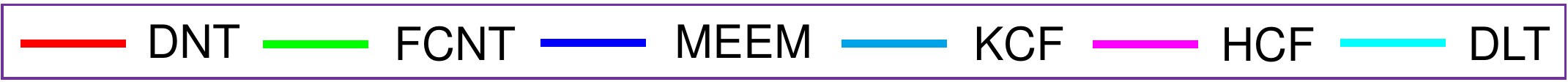}
	\end{tabular}
	\begin{tabular}{c@{}c@{}c@{}c@{}c@{}c@{}c@{}c}
\hspace{-3mm}
		\includegraphics[width=0.12\linewidth,height=1.4cm]{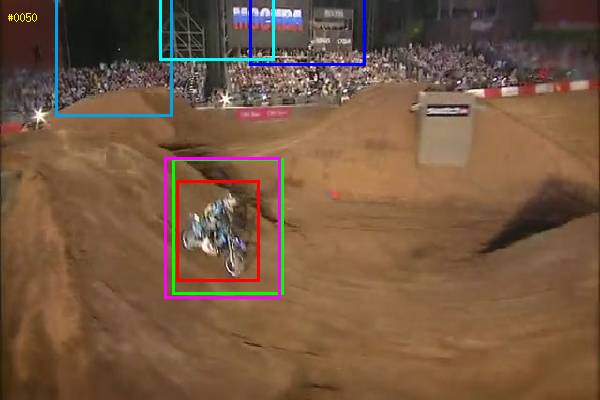}
		\ &
		\includegraphics[width=0.12\linewidth,height=1.4cm]{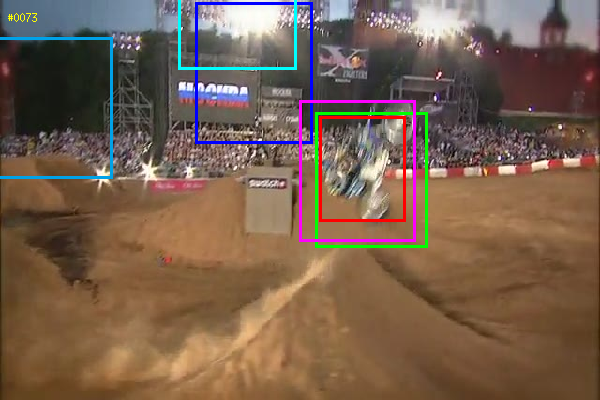}
		\ &
		\includegraphics[width=0.12\linewidth,height=1.4cm]{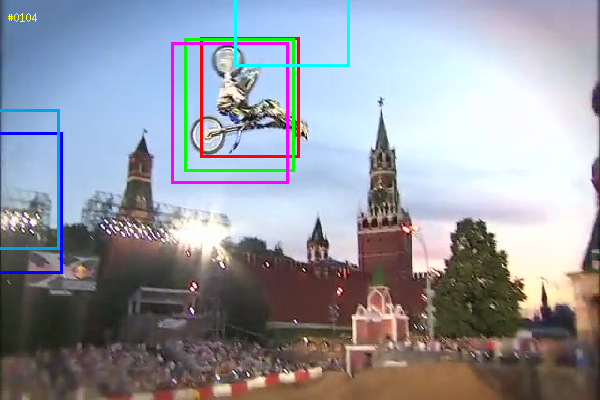}
		\ &
		\includegraphics[width=0.12\linewidth,height=1.4cm]{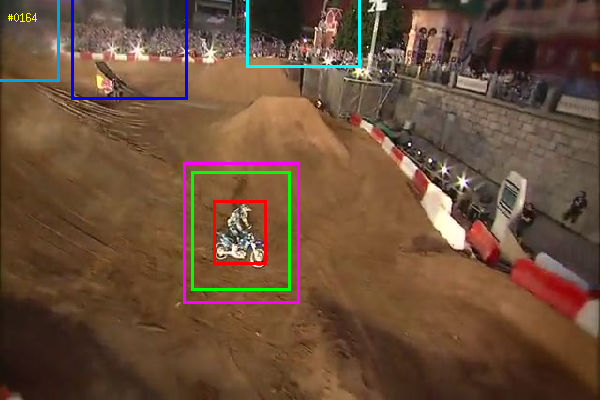}
		\ &
		\includegraphics[width=0.12\linewidth,height=1.4cm]{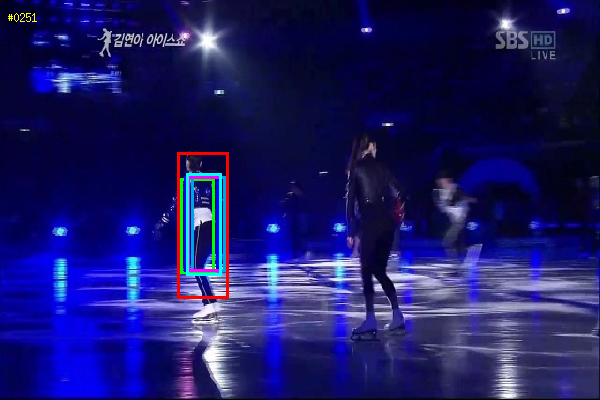}
		\ &
		\includegraphics[width=0.12\linewidth,height=1.4cm]{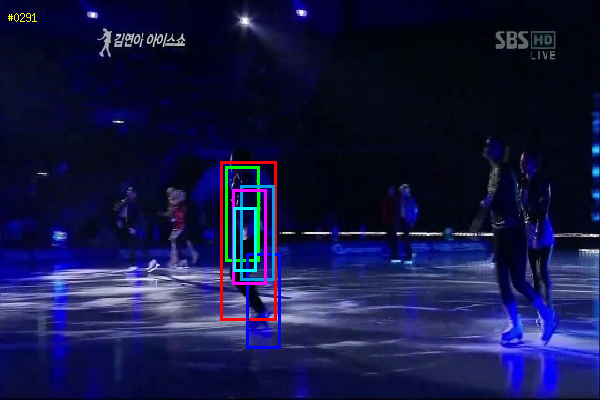}
		\ &
		\includegraphics[width=0.12\linewidth,height=1.4cm]{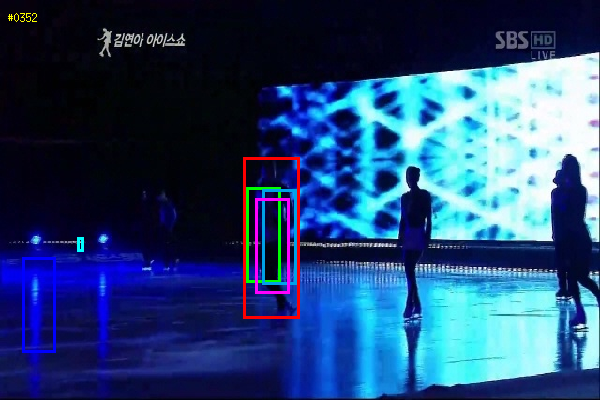}
		\ &
		\includegraphics[width=0.12\linewidth,height=1.4cm]{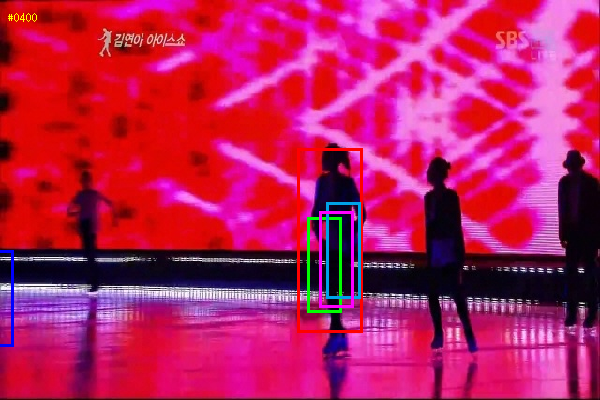}
	\end{tabular}
	\\
	\begin{tabular}{c@{}c@{}c@{}c@{}c@{}c@{}c@{}c}
\hspace{-3mm}
		\includegraphics[width=0.12\linewidth,height=1.4cm]{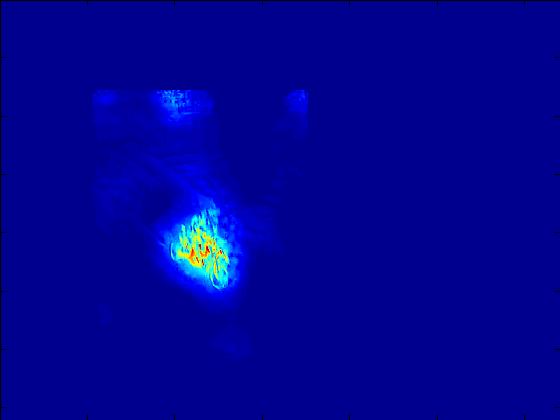}
		\ &
		\includegraphics[width=0.12\linewidth,height=1.4cm]{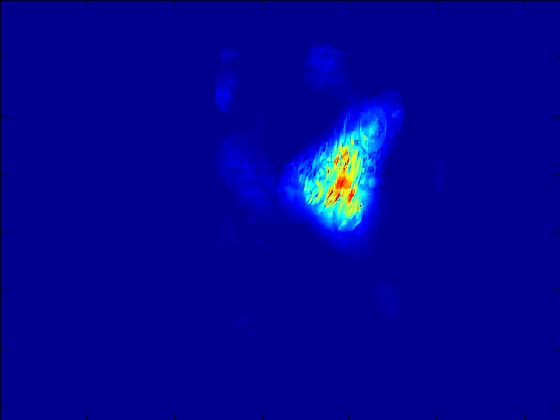}
		\ &
		\includegraphics[width=0.12\linewidth,height=1.4cm]{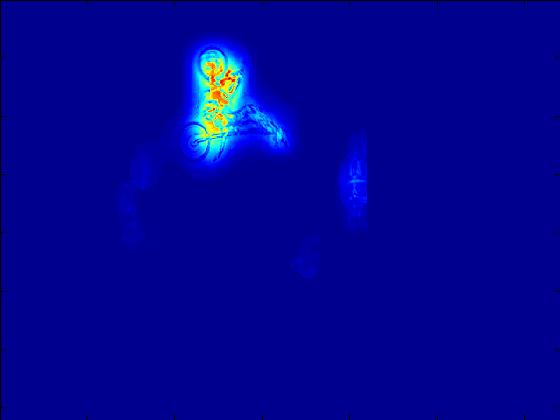}
		\ &
		\includegraphics[width=0.12\linewidth,height=1.4cm]{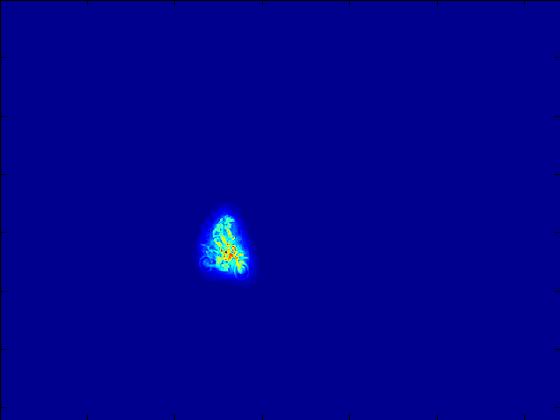}
		\ &
		\includegraphics[width=0.12\linewidth,height=1.4cm]{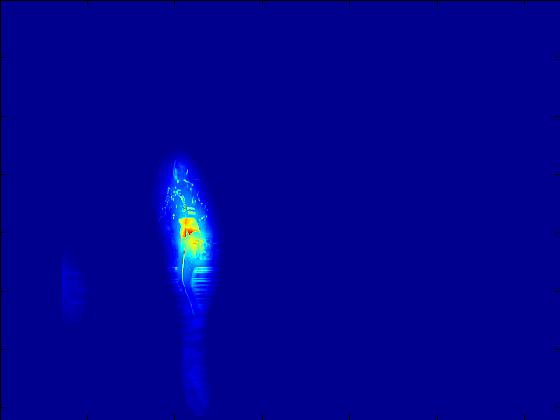}
		\ &
		\includegraphics[width=0.12\linewidth,height=1.4cm]{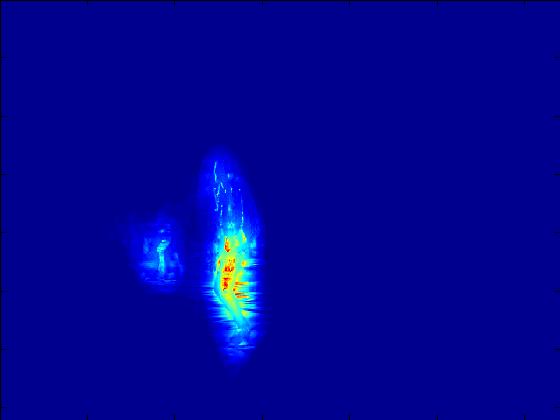}
		\ &
		\includegraphics[width=0.12\linewidth,height=1.4cm]{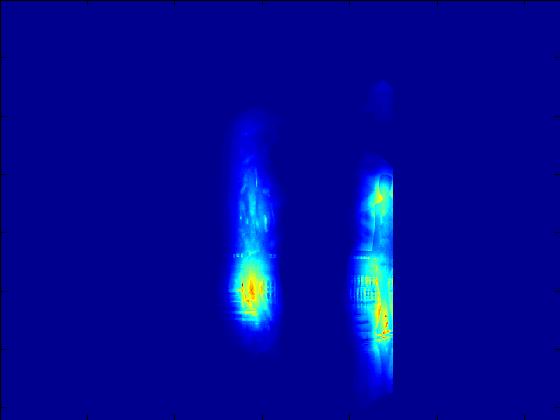}
		\ &
		\includegraphics[width=0.12\linewidth,height=1.4cm]{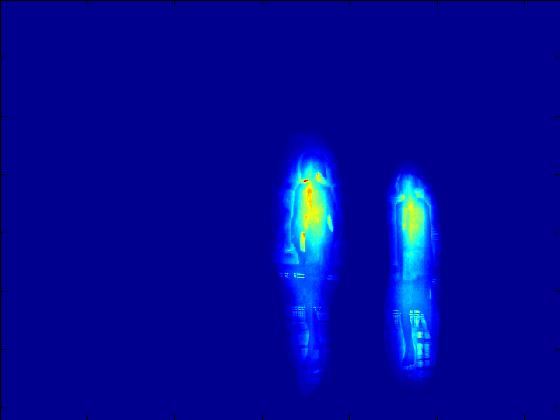}
	\end{tabular}
	\\
	\begin{tabular}{c@{}c@{}c@{}c@{}c@{}c@{}c@{}c}
\hspace{-3mm}
		\includegraphics[width=0.12\linewidth,height=1.4cm]{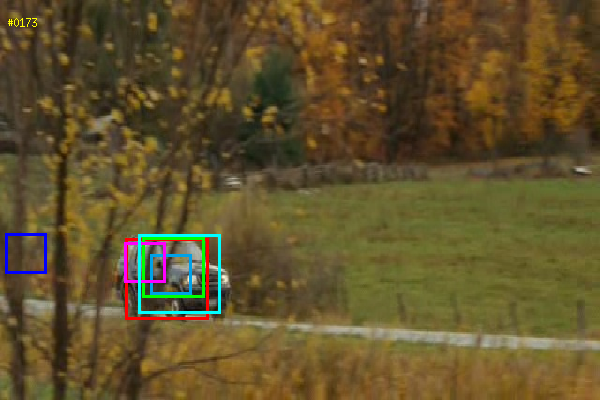}
		\ &
		\includegraphics[width=0.12\linewidth,height=1.4cm]{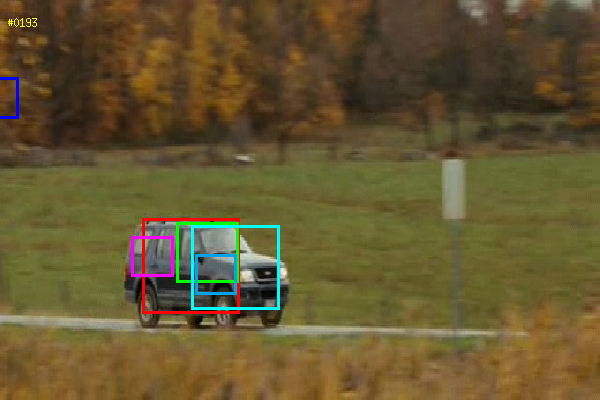}
		\ &
		\includegraphics[width=0.12\linewidth,height=1.4cm]{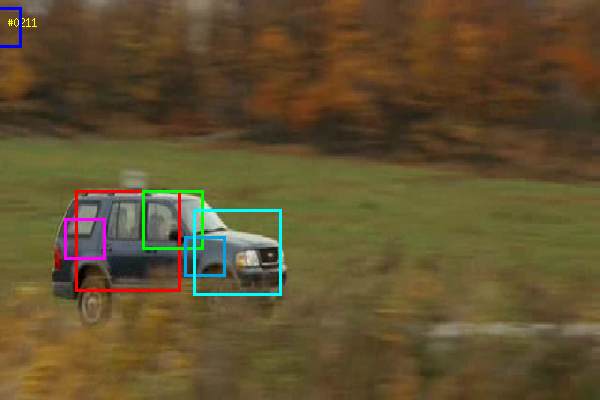}
		\ &
		\includegraphics[width=0.12\linewidth,height=1.4cm]{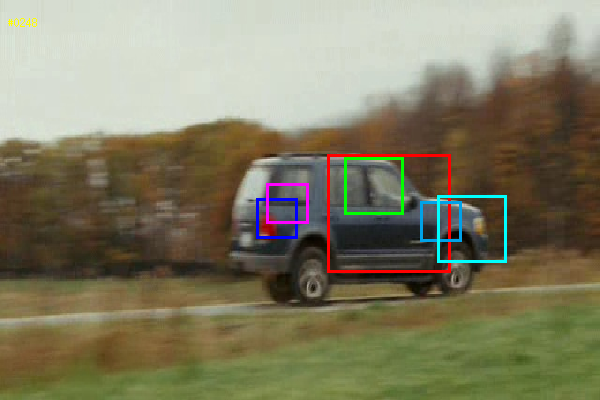}
		\ &
		\includegraphics[width=0.12\linewidth,height=1.4cm]{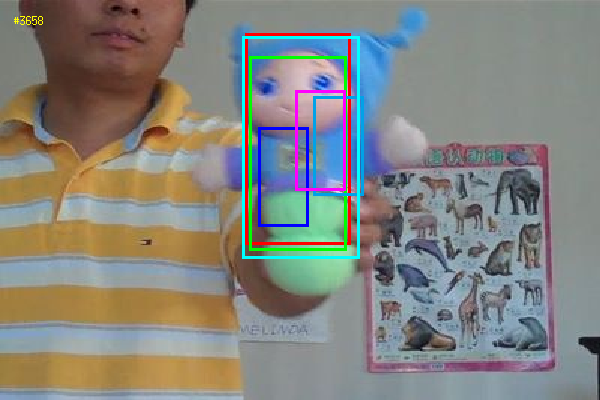}
		\ &
		\includegraphics[width=0.12\linewidth,height=1.4cm]{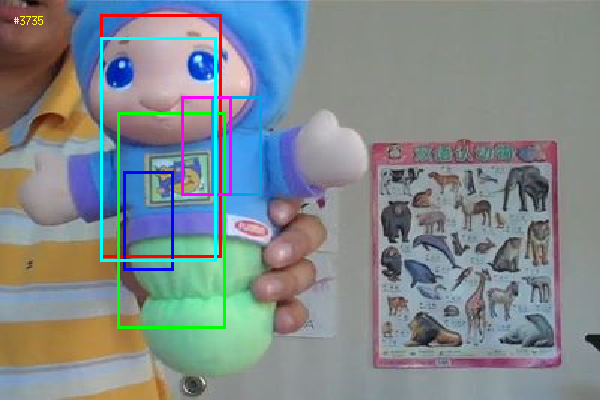}
		\ &
		\includegraphics[width=0.12\linewidth,height=1.4cm]{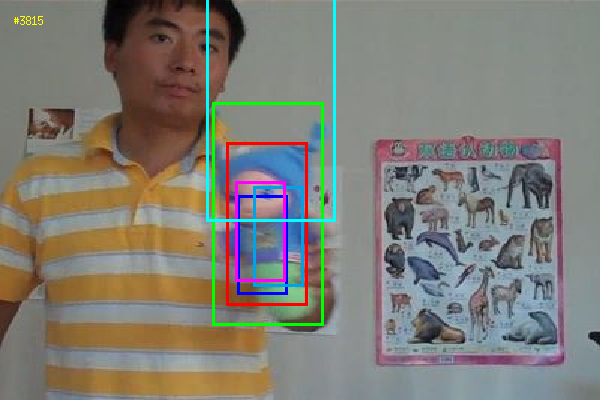}
		\ &
		\includegraphics[width=0.12\linewidth,height=1.4cm]{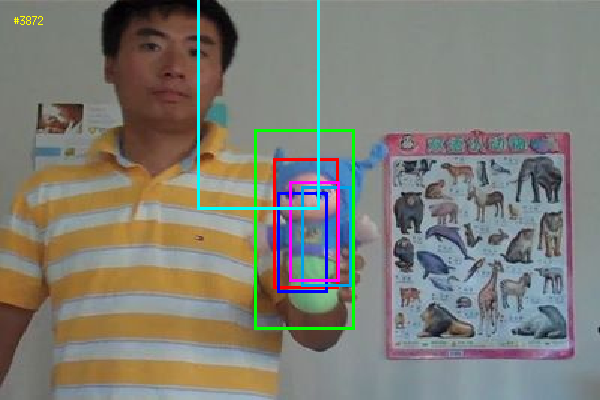}
	\end{tabular}
	\\
	\begin{tabular}{c@{}c@{}c@{}c@{}c@{}c@{}c@{}c}
\hspace{-3mm}
		\includegraphics[width=0.12\linewidth,height=1.4cm]{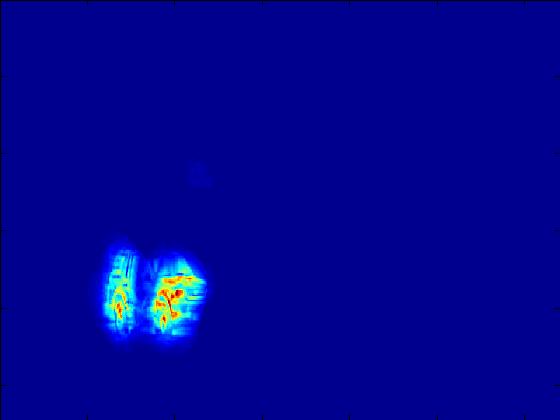}
		\ &
		\includegraphics[width=0.12\linewidth,height=1.4cm]{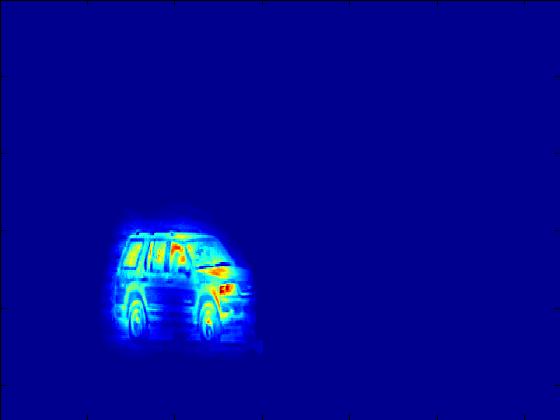}
		\ &
		\includegraphics[width=0.12\linewidth,height=1.4cm]{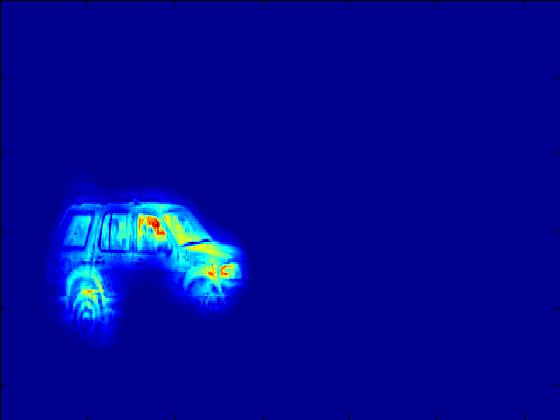}
		\ &
		\includegraphics[width=0.12\linewidth,height=1.4cm]{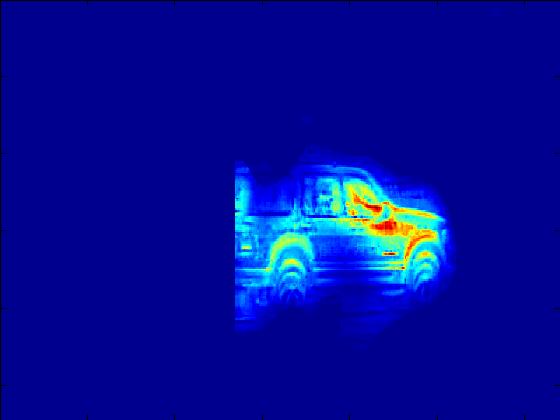}
		\ &
		\includegraphics[width=0.12\linewidth,height=1.4cm]{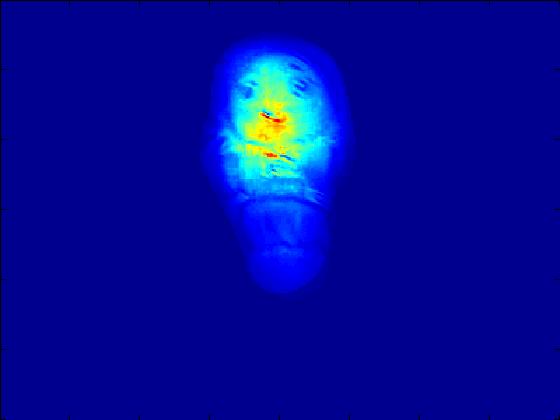}
		\ &
		\includegraphics[width=0.12\linewidth,height=1.4cm]{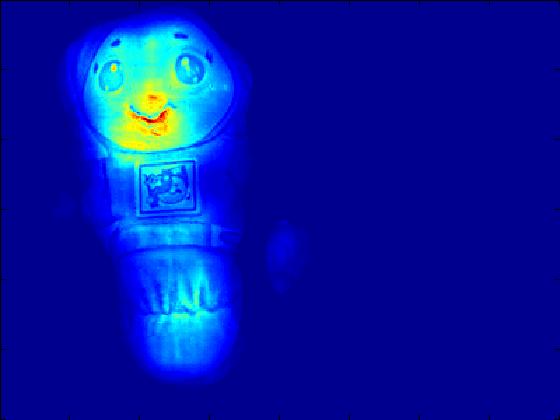}
		\ &
		\includegraphics[width=0.12\linewidth,height=1.4cm]{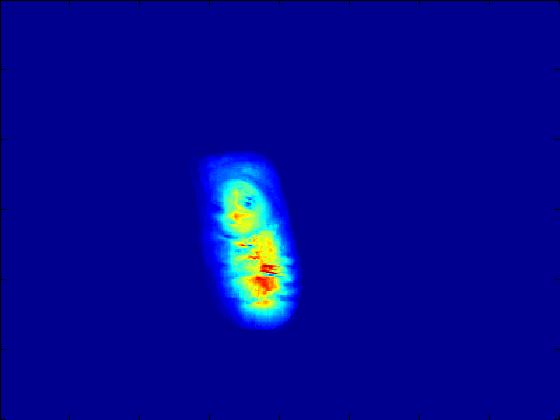}
		\ &
		\includegraphics[width=0.12\linewidth,height=1.4cm]{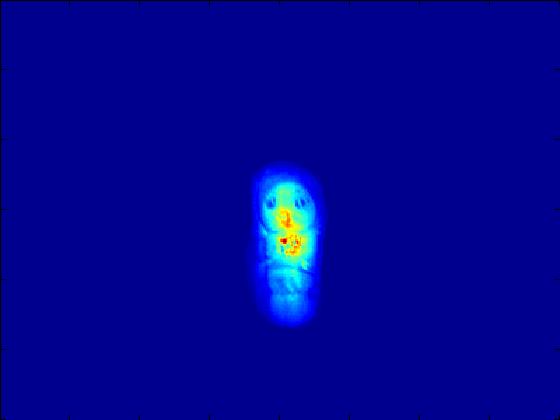}
	\end{tabular}
	\\
    \begin{tabular}{c@{}c@{}c@{}c@{}c@{}c@{}c@{}c}
    \hspace{-3mm}
		\includegraphics[width=0.12\linewidth,height=1.4cm]{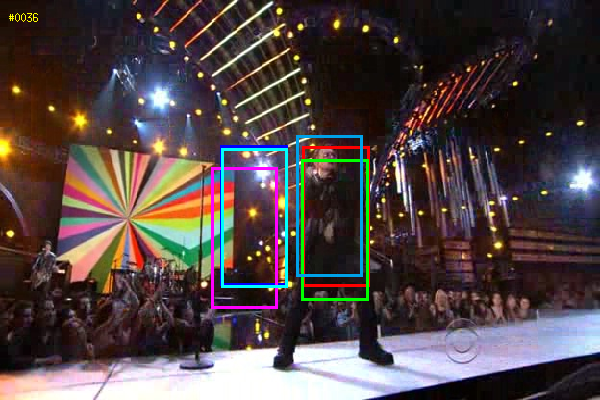}
		\ &
		\includegraphics[width=0.12\linewidth,height=1.4cm]{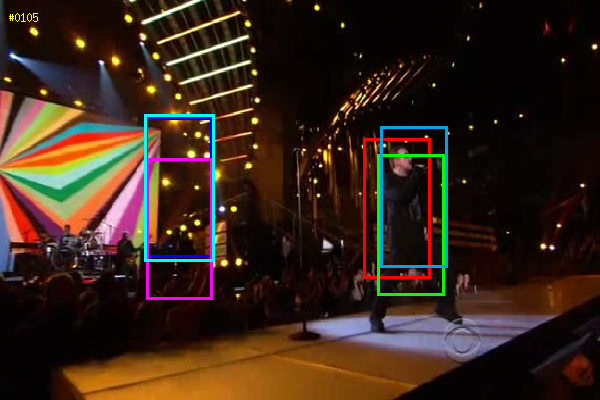}
		\ &
		\includegraphics[width=0.12\linewidth,height=1.4cm]{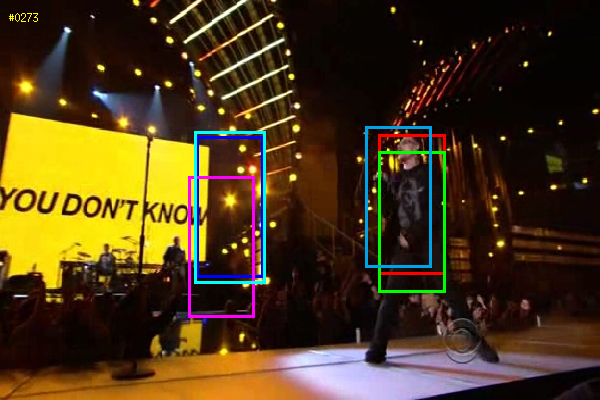}
		\ &
		\includegraphics[width=0.12\linewidth,height=1.4cm]{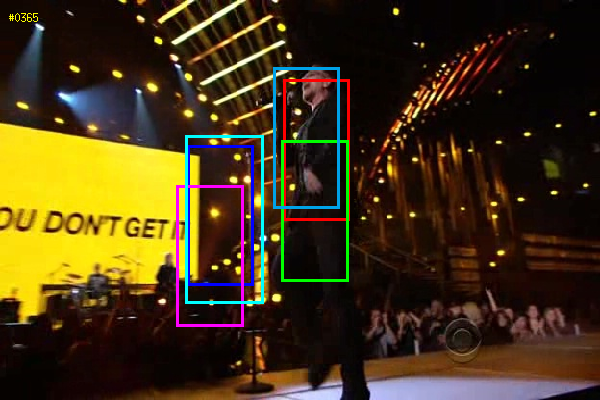}
		\ &
		\includegraphics[width=0.12\linewidth,height=1.4cm]{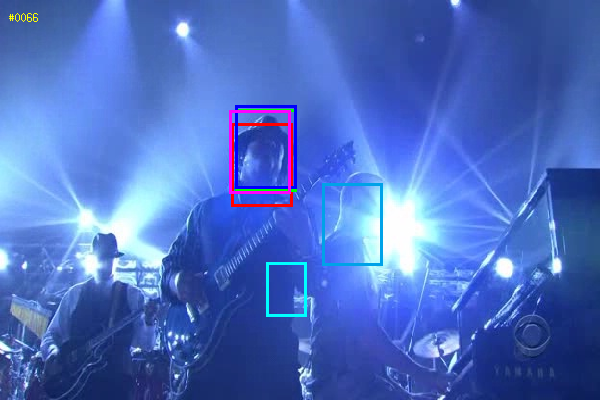}
		\ &
		\includegraphics[width=0.12\linewidth,height=1.4cm]{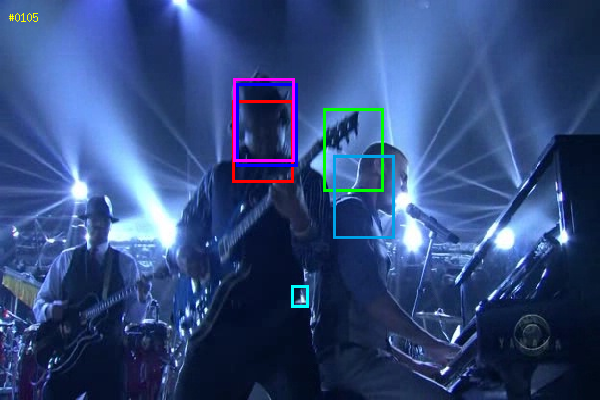}
		\ &
		\includegraphics[width=0.12\linewidth,height=1.4cm]{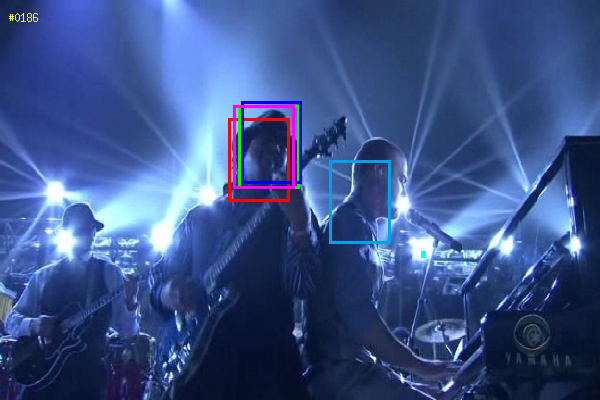}
		\ &
		\includegraphics[width=0.12\linewidth,height=1.4cm]{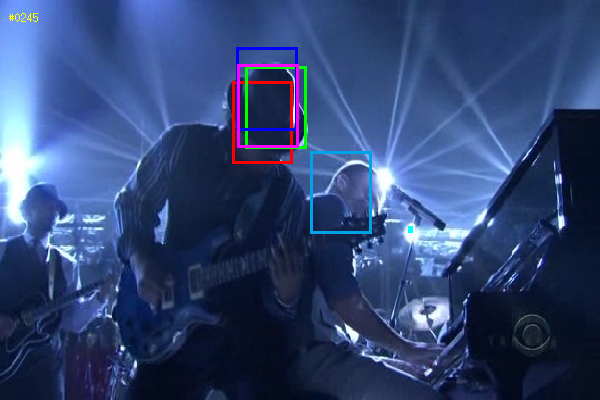}
	\end{tabular}
	\\
    \begin{tabular}{c@{}c@{}c@{}c@{}c@{}c@{}c@{}c}
    \hspace{-3mm}
		\includegraphics[width=0.12\linewidth,height=1.4cm]{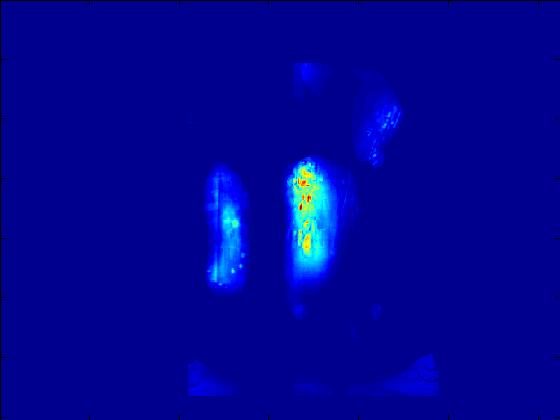}
		\ &
		\includegraphics[width=0.12\linewidth,height=1.4cm]{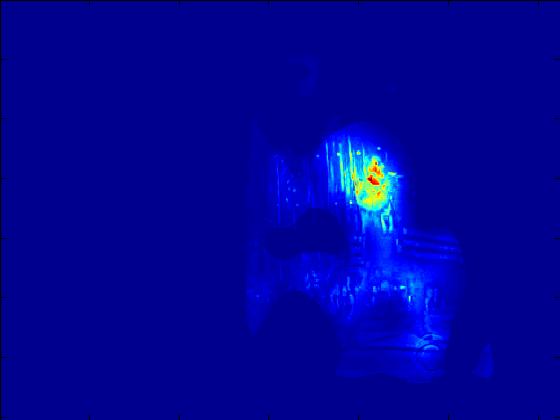}
		\ &
		\includegraphics[width=0.12\linewidth,height=1.4cm]{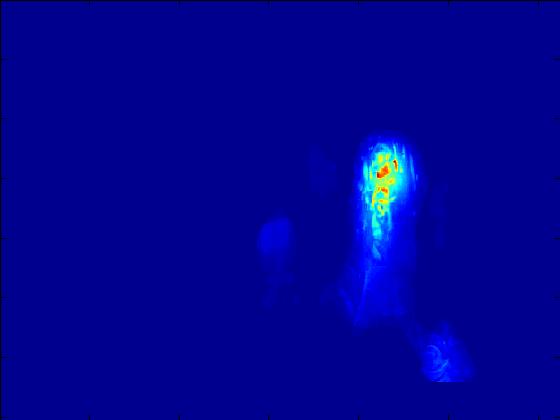}
		\ &
		\includegraphics[width=0.12\linewidth,height=1.4cm]{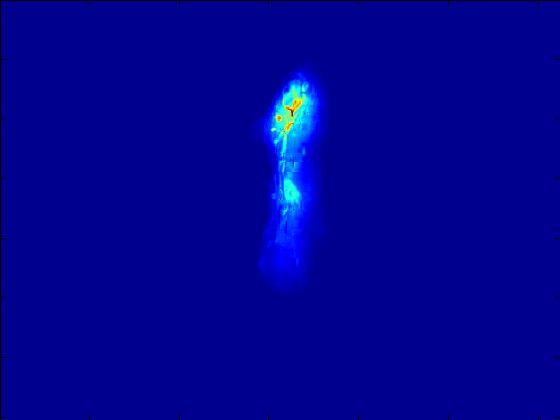}
		\ &
		\includegraphics[width=0.12\linewidth,height=1.4cm]{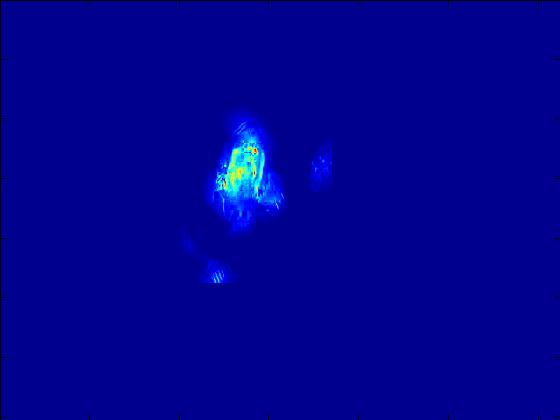}
		\ &
		\includegraphics[width=0.12\linewidth,height=1.4cm]{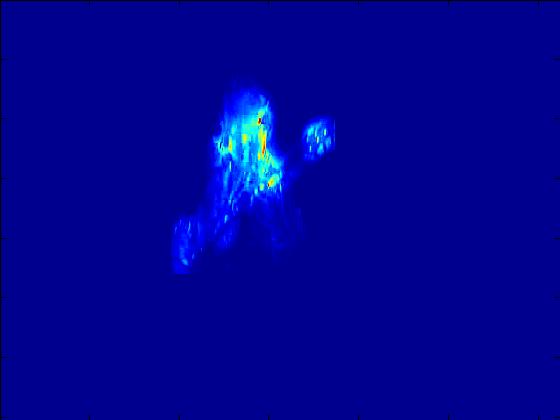}
		\ &
		\includegraphics[width=0.12\linewidth,height=1.4cm]{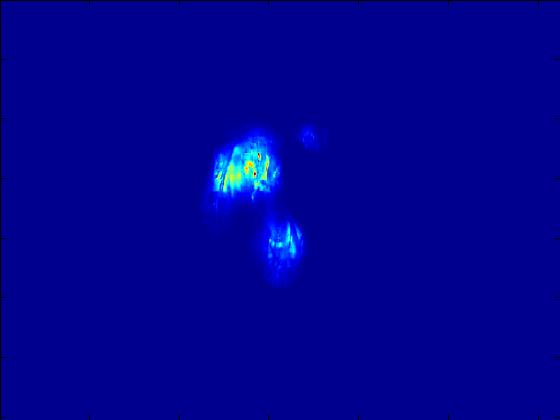}
		\ &
		\includegraphics[width=0.12\linewidth,height=1.4cm]{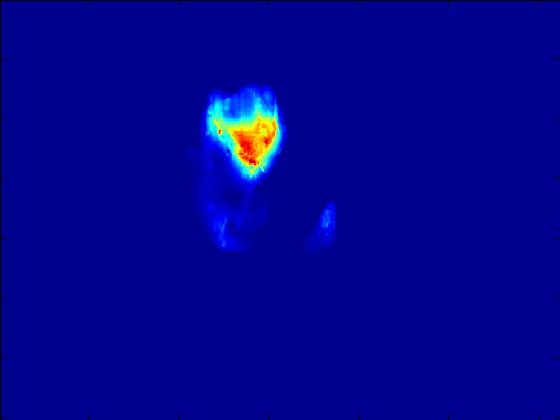}
	\end{tabular}
	\\
	\vspace{-1mm}
	\caption{Qualitative evaluation of our algorithm \textcolor[rgb]{1,0,0}{DNT}, \textcolor[rgb]{0,1,0}{FCNT}~\cite{wangvisual},
		\textcolor[rgb]{0,0,1}{MEEM}~\cite{zhang2014meem}, \textcolor[rgb]{0,0.63529,0.9098}{KCF}~\cite{henriques15},
		\textcolor[rgb]{1,0,1}{HCF}~\cite{machao15}, \textcolor[rgb]{0,1,1}{DLT}~\cite{wang13}. Six challenging sequences (from left to right and top to down are \emph{MotorRolling}, \emph{Skating}, \emph{CarScale}, \emph{Doll}, \emph{Singer2}, \emph{Shaking}, respectively)
		are compared among the state-of-the-art trackers. Odd row: Comparisons with state-of-the-art trackers. Even row: Combined ICA-R maps with a weighted coefficient.}
	\label{qualitative_results}
	\vspace{-1mm}
\end{figure*}

{\flushleft{\textbf{Qualitative Evaluation.}}}
Figure~\ref{qualitative_results}
shows some visual results compared with other state-of-the-art trackers.
%
The even rows are the combined ICA-R maps $\textbf{v}^k,k=\{1,2\}$ with a weighted coefficient
$\lambda$ as~\eqref{maxconfidence} for target localization.
The semantic information and clear outline of the tracking targets are highlighted with the help of ConvNet features and boundary guidance.
It is worth mentioning
that in the most challenging sequences such as \emph{MotorRolling} and \emph{Skating} (row 1), most methods fail to track targets well whereas our
DNT algorithm performs accurately in terms of either precision or overlap;
%
%
For the most representative sequences for scaling (row 3),
other trackers are capable to follow a part of target object, but drift away when the target
objects undergo large scale variation.
Despite these challenges, the DNT algorithm accurately estimates
both scales and positions of target objects;

For the sequences (row 5) to deal with weak network activations, the target objects are similar to background in appearance. Even though semantics cannot be highlighted in the higher layers, the fine-grained spatial details are kept for accurate localization;


Among these sequences, despite fast motion and significant distractors, the proposed
algorithm can still track the target object with the help of high-level features of
targets from the dual network.

\subsection{Evaluation on VOT2015}

{\flushleft{\textbf{Data Set and Evaluation Settings.}}}
The VOT2015 data set~\cite{kristan2015visual} contains 60 video sequences showing various objects in challenging background.
The sequences are chosen from a large pool of real-life data sets. According to the evaluation criterion, the proposed tracker is re-initialized with the ground truth location whenever tracking fails (the overlap between the estimated location and ground truth location equals to zero). Following~\cite{kristan2015visual}, we conduct two experiments: baseline and Overall evaluation, the trackers are initialized with the ground truth location. Two metrics are used to rank all the trackers: accuracy and robustness. Accuracy measures the overlap ratio between the estimated and ground truth bounding box, while robustness measures the probability of tracking failures. We evaluate the proposed DNT tracker with all the trackers submitted to VOT2015 challenge~\cite{kristan2015visual}.
More details about the evaluation protocol and compared trackers can be found in~\cite{kristan2015visual}.

{\flushleft{\textbf{Evaluation Results.}}}
Considering the limited space, we only present the average accuracy and robustness rank of the top twelve trackers in Table~\ref{vot2015}. In general, the proposed DNT tracker ranks the 3rd in the VOT2015 challenge~\cite{kristan2015visual}. In both the baseline and Overall experiments, DNT performs well compared with other state-of-the-art trackers. We mainly fail in some sequences such as \emph{fish1-4}, \emph{leaves}, \emph{octopus} \emph{et al}. Because the target is quite small and similar to surrounding distractors, the dual net can hardly achieve the semantic information, which leads to drifting easily. In such cases, the features on the lower layers of CNN model are able to track target better since fine-grained spatial information weighs more.
For the \emph{soldier} and \emph{sheep} sequences, when long-term occlusions happen, the proposed method fails to learn the appearance model well since there is no additional detection mechanism.

Figure~\ref{vot_fig} shows the accuracy and robustness plots for the top 24 trackers in the VOT2015 data set~\cite{kristan2015visual}. The best tracker is located at the upper-right corner. From the figures we conclude that the DNT method performs favorably against the other methods.

\subsection{Ablation Study}\label{ablationstudy}

First, we demonstrate the effectiveness of boundary information, ICA-R, random samples and the stochastic update mechanism in training and updating the network.
We use the DNT tracker as the baseline comparison where the feature representation descends from \texttt{conv4-3} and \texttt{conv5-3} of the VGG network in the OTB50 date set~\cite{wu2013online}.
We compare the proposed DNT algorithm with variants without boundary information, without ICA-R, without updating random samples and
without stochastic update.
Table~\ref{ablation} shows that the full functionality DNT algorithm with all the components performs much better than other variants.
We have also analysed the importance of boundary information and random update mechanism on different tracking challenges.
Table~\ref{ablation3} shows the average precision and success scores on different attributes in the OPE experiment for OTB50~\cite{wu2013online}.
DNT\textsubscript{B} and DNT\textsubscript{R} denote two variations of the DNT tracker without boundary reference and
without random sample update.

\begin{table}[t]
	\centering
	\caption{Ablation study of different components of the DNT tracker. The precision and overlap are reported on the benchmark data set \cite{wu2013online}.}
	\label{ablation}
	\footnotesize{
		\begin{tabular}{c@{}c}
        \includegraphics[width=0.6\linewidth, height=0.3\linewidth]{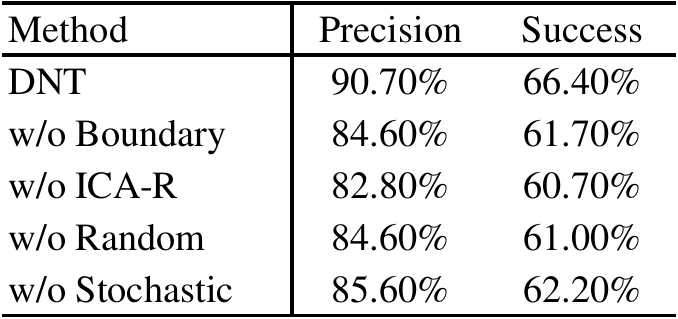} \\
		\end{tabular}
	}
\vspace{-0.5cm}
\end{table}

\begin{table}[t]
	\centering
	\caption{Ablation study of different strategies under different attributes of the DNT tracker. $\text{DNT}_\text{B}$ denotes without boundary guidance and $\text{DNT}_\text{R}$ indicates without random update. The precision and overlap are reported on the benchmark data set~\cite{wu2013online}.}
	\label{ablation3}
	\footnotesize{
		\begin{tabular}{c@{}c}
        \includegraphics[width=\linewidth, height=0.18\linewidth]{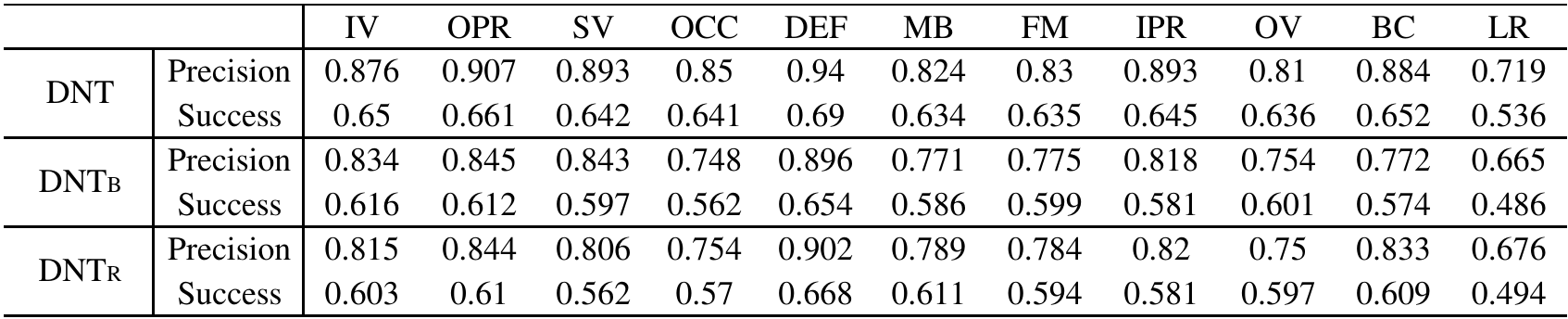} \\
		\end{tabular}
	}
\vspace{-0.2cm}
\end{table}

\begin{table}[t]
	\centering
	\footnotesize
	\caption{Ablation study of feature combination among layers.
	The precision and overlap are reported on the benchmark data set~\cite{wu2013online}.}
	\label{ablation2}
    \begin{tabular}{c@{}c}
    \includegraphics[width=\linewidth]{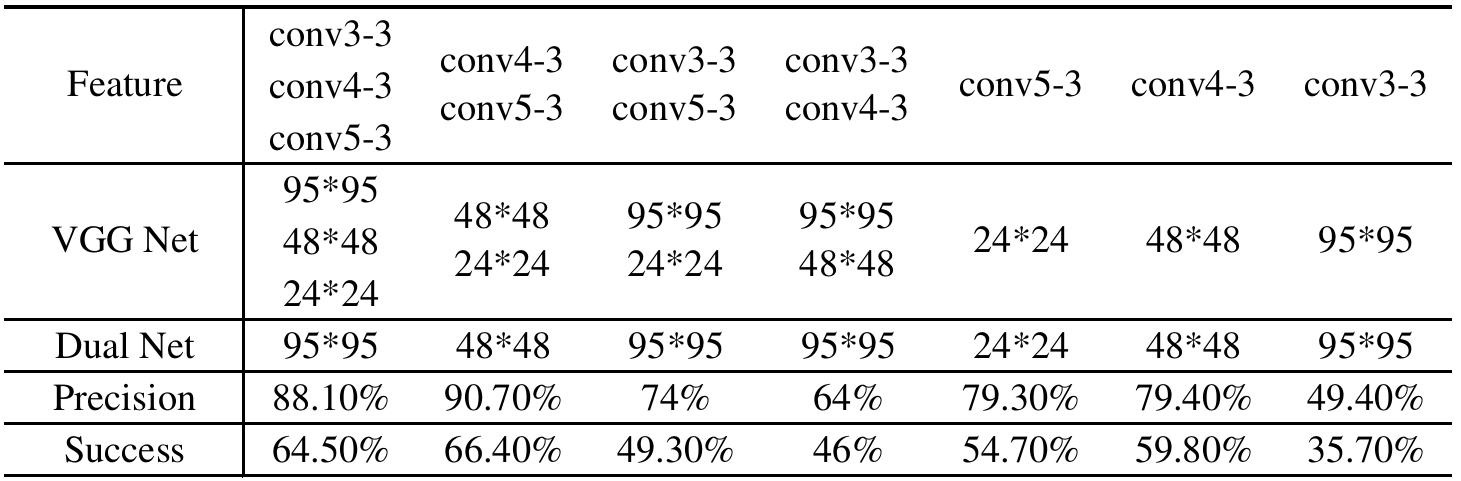} \\
    \end{tabular}
    \vspace{-0.4cm}
\end{table}

Second, we analyze the feature selection from different layers to represent target objects
(Table \ref{ablation2}).
The reason as for using \texttt{conv4-3} and \texttt{conv5-3} as the feature extraction layers
 is based on the empirical analysis that clear outlines of target objects can be obtained.
Table \ref{ablation2} shows that the tracker using a dual layer outperforms
that with one single layer because both semantic and fine-grained cues are considered.
We have also conducted the experiment with \texttt{conv3-3}, \texttt{conv4-3} and \texttt{conv5-3} features, $\lambda$ in~\eqref{maxconfidence} is set [0.2, 0.3, 0.5] for each layer. The performance is not so good as our base tracker because the outline of target object can hardly be reserved in \texttt{conv3-3}.
In addition, the trackers using features from higher layers perform better than those using
features from lower layers, which shows
that semantic appearance plays an important role for visual tracking.




\section{Conclusion}
In this paper, we propose an effective hierarchical feature learning algorithm for visual tracking. We offline learn powerful ConvNet features from the VGG network on a large-scale data set.
Both higher layer features with semantics and lower layers with spatial structure are used for visual tracking.
Integrated with boundary information, the two streams of features are able to delineate a coarse prior map of the target object.
These features serve as reference to extract better target appearance model by the ICA-R method.
The dual network is trained and updated as an adaptation module to fine-tune
the ConvNet features for specific target object.
We put forward a course-to-fine strategy to locate target and a self-supervised scheme to update the dual network.
Massive quantitative and qualitative experiments against the state-of-the-art algorithms based on OTB and VOT2015 data sets demonstrate the accuracy and robustness of the proposed method.


\bibliographystyle{IEEEtran}
\bibliography{ref_tracking}

\begin{thebibliography}{10}
\providecommand{\url}[1]{#1}
\csname url@samestyle\endcsname
\providecommand{\newblock}{\relax}
\providecommand{\bibinfo}[2]{#2}
\providecommand{\BIBentrySTDinterwordspacing}{\spaceskip=0pt\relax}
\providecommand{\BIBentryALTinterwordstretchfactor}{4}
\providecommand{\BIBentryALTinterwordspacing}{\spaceskip=\fontdimen2\font plus
\BIBentryALTinterwordstretchfactor\fontdimen3\font minus
  \fontdimen4\font\relax}
\providecommand{\BIBforeignlanguage}[2]{{%
\expandafter\ifx\csname l@#1\endcsname\relax
\typeout{** WARNING: IEEEtran.bst: No hyphenation pattern has been}%
\typeout{** loaded for the language `#1'. Using the pattern for}%
\typeout{** the default language instead.}%
\else
\language=\csname l@#1\endcsname
\fi
#2}}
\providecommand{\BIBdecl}{\relax}
\BIBdecl

\bibitem{ren07}
X.~Ren and J.~Malik, ``Tracking as repeated figure/ground segmentation,'' in
  \emph{CVPR}, 2007, pp. 1--8.

\bibitem{ross2008}
D.~A. Ross, J.~Lim, R.-S. Lin, and M.-H. Yang, ``Incremental learning for
  robust visual tracking,'' \emph{IJCV}, vol.~77, no. 1-3, pp. 125--141, 2008.

\bibitem{grabner08}
H.~Grabner, C.~Leistner, and H.~Bischof, ``Semi-supervised on-line boosting for
  robust tracking,'' in \emph{ECCV}, 2008, pp. 234--247.

\bibitem{mei09}
X.~Mei and H.~Ling, ``Robust visual tracking using l1 minimization,'' in
  \emph{CVPR}, 2009, pp. 1436--1443.

\bibitem{hare2011}
S.~Hare, A.~Saffari, and P.~H. Torr, ``Struck: Structured output tracking with
  kernels,'' in \emph{ICCV}, 2011, pp. 263--270.

\bibitem{jia2012visual}
X.~Jia, H.~Lu, and M.-H. Yang, ``Visual tracking via adaptive structural local
  sparse appearance model,'' in \emph{CVPR}, 2012, pp. 1822--1829.

\bibitem{zhong14}
W.~Zhong, H.~Lu, and M.~Yang, ``Robust object tracking via sparse collaborative
  appearance model,'' \emph{TIP}, vol.~23, no.~5, pp. 2356--2368, 2014.

\bibitem{zhang2014meem}
J.~Zhang, S.~Ma, and S.~Sclaroff, ``Meem: Robust tracking via multiple experts
  using entropy minimization,'' in \emph{ECCV}, 2014, pp. 188--203.

\bibitem{krizhevsky12}
A.~Krizhevsky, I.~Sutskever, and G.~E. Hinton, ``Imagenet classification with
  deep convolutional neural networks,'' in \emph{NIPS}, 2012, pp. 1106--1114.

\bibitem{li2016multi}
H.~Li, W.~Ouyang, and X.~Wang, ``Multi-bias non-linear activation in deep
  neural networks,'' in \emph{International Conference on Machine Learning},
  2016.

\bibitem{simonyan2014very}
K.~Simonyan and A.~Zisserman, ``Very deep convolutional networks for
  large-scale image recognition,'' \emph{arXiv preprint: 1409.1556}, 2014.

\bibitem{he2015deep}
K.~He, X.~Zhang, S.~Ren, and J.~Sun, ``Deep residual learning for image
  recognition,'' \emph{CVPR}, 2016.

\bibitem{girshick14}
R.~Girshick, J.~Donahue, T.~Darrell, and J.~Malik, ``Rich feature hierarchies
  for accurate object detection and semantic segmentation,'' in \emph{CVPR},
  2014, pp. 580--587.

\bibitem{girshick2015fast}
R.~Girshick, ``Fast r-cnn,'' in \emph{ICCV}, 2015.

\bibitem{li2015inner}
H.~Li, H.~Lu, Z.~Lin, X.~Shen, and B.~Price, ``Inner and inter label
  propagation: Salient object detection in the wild,'' \emph{IEEE Transactions
  on Image Processing}, vol.~24, no.~10, pp. 3176--3186, 2015.

\bibitem{li2016cnn_sal}
H.~Li, J.~Chen, H.~Lu, and Z.~Chi, ``{CNN} for saliency detection with
  low-level feature integration,'' \emph{Elsevier Neurocomputing}, 2016.

\bibitem{wang2012}
Q.~Wang, F.~Chen, J.~Yang, W.~Xu, and M.-H. Yang, ``Transferring visual prior
  for online object tracking,'' \emph{TIP}, vol.~21, no.~7, pp. 3296--3305,
  2012.

\bibitem{wang13}
N.~Wang and D.~Yeung, ``Learning a deep compact image representation for visual
  tracking,'' in \emph{NIPS}, 2013, pp. 809--817.

\bibitem{hong2015tracking}
S.~Hong, T.~You, S.~Kwak, and B.~Han, ``Online tracking by learning
  discriminative saliency map with convolutional neural network,'' in
  \emph{ICML}, 2015.

\bibitem{li_deeptrack}
H.~Li, Y.~Li, and F.~Porikli, ``Deeptrack: Learning discriminative feature
  representations by convolutional neural networks for visual tracking,'' in
  \emph{BMVC}, 2014.

\bibitem{wang2015video}
L.~Wang, T.~Liu, G.~Wang, K.~L. Chan, and Q.~Yang, ``Video tracking using
  learned hierarchical features,'' \emph{TIP}, vol.~24, no.~4, pp. 1424--1435,
  2015.

\bibitem{nam2015learning}
H.~Nam and B.~Han, ``Learning multi-domain convolutional neural networks for
  visual tracking,'' \emph{CVPR}, 2016.

\bibitem{bappy2016cnn}
J.~H. Bappy and A.~K. Roy-Chowdhury, ``Cnn based region proposals for efficient
  object detection,'' in \emph{ICIP}, 2016, pp. 3658--3662.

\bibitem{hypercolumn}
B.~Hariharan, P.~Arbeláez, R.~Girshick, and J.~Malik, ``Hypercolumns for
  object segmentation and fine-grained localization,'' in \emph{CVPR}, 2014.

\bibitem{LuR06}
W.~Lu and J.~C. Rajapakse, ``{ICA} with reference,'' \emph{Neurocomputing},
  vol.~69, no. 16-18, pp. 2244--2257, 2006.

\bibitem{wu2013online}
Y.~Wu, J.~Lim, and M.-H. Yang, ``Onine object tracking: A benchmark,'' in
  \emph{CVPR}.\hskip 1em plus 0.5em minus 0.4em\relax IEEE, 2013, pp.
  2411--2418.

\bibitem{kristan2015visual}
M.~Kristan, J.~Matas, A.~Leonardis, M.~Felsberg, L.~Cehovin, G.~Fernandez,
  T.~Vojir, G.~Hager, G.~Nebehay, and R.~Pflugfelder, ``The visual object
  tracking vot2015 challenge results,'' in \emph{ECCV Workshops}, 2015.

\bibitem{wu2015object}
Y.~Wu, J.~Lim, and M.-H. Yang, ``Object tracking benchmark,'' \emph{TPAMI},
  vol.~37, no.~9, pp. 1834--1848, 2015.

\bibitem{avidan2004support}
S.~Avidan, ``Support vector tracking,'' \emph{TPAMI}, vol.~26, no.~8, pp.
  1064--1072, 2004.

\bibitem{babenko11}
B.~Babenko, M.-H. Yang, and S.~Belongie, ``Robust object tracking with online
  multiple instance learning,'' \emph{TPAMI}, vol.~33, no.~8, pp. 1619--1632,
  2011.

\bibitem{CSK12}
J.~F. Henriques, R.~Caseiro, P.~Martins, and J.~Batista, ``Exploiting the
  circulant structure of tracking-by-detection with kernels,'' in \emph{ECCV},
  2012, pp. 702--715.

\bibitem{bolme2010visual}
D.~S. Bolme, J.~R. Beveridge, B.~Draper, Y.~M. Lui \emph{et~al.}, ``Visual
  object tracking using adaptive correlation filters,'' in \emph{CVPR}, 2010,
  pp. 2544--2550.

\bibitem{danelljan2014accurate}
M.~Danelljan, G.~H{\"a}ger, F.~Khan, and M.~Felsberg, ``Accurate scale
  estimation for robust visual tracking,'' in \emph{BMVC}, 2014.

\bibitem{henriques15}
J.~F. Henriques, R.~Caseiro, P.~Martins, and J.~Batista, ``High-speed tracking
  with kernelized correlation filters,'' \emph{TPAMI}, vol.~37, no.~3, pp.
  583--596, 2015.

\bibitem{danelljan2014adaptive}
M.~Danelljan, F.~S. Khan, M.~Felsberg, and J.~van~de Weijer, ``Adaptive color
  attributes for real-time visual tracking,'' in \emph{CVPR}, 2014, pp.
  1090--1097.

\bibitem{wangvisual}
L.~Wang, W.~Ouyang, X.~Wang, and H.~Lu, ``Visual tracking with fully
  convolutional networks,'' in \emph{ICCV}, 2015, pp. 3119--3127.

\bibitem{machao15}
C.~Ma, J.-B. Huang, X.~Yang, and M.-H. Yang, ``Hierarchical convolutional
  features for visual tracking,'' in \emph{ICCV}, 2015, pp. 3074--3082.

\bibitem{ondruska2016deep}
P.~Ondruska and I.~Posner, ``Deep tracking: Seeing beyond seeing using
  recurrent neural networks,'' \emph{arXiv preprint:1602.00991}, 2016.

\bibitem{cui2016recurrently}
Z.~Cui, S.~Xiao, J.~Feng, and S.~Yan, ``Recurrently target-attending
  tracking,'' in \emph{CVPR}, 2016, pp. 1449--1458.

\bibitem{marr1980theory}
D.~Marr and E.~Hildreth, ``Theory of edge detection,'' \emph{Proceedings of the
  Royal Society of London B: Biological Sciences}, vol. 207, no. 1167, pp.
  187--217, 1980.

\bibitem{xiaolong}
X.~Wang and A.~Gupta, ``Unsupervised learning of visual representations using
  videos,'' in \emph{ICCV}, 2015, pp. 2794--2802.

\bibitem{jia14}
Y.~Jia, E.~Shelhamer, J.~Donahue, S.~Karayev, J.~Long, R.~Girshick,
  S.~Guadarrama, and T.~Darrell, ``Caffe: Convolutional architecture for fast
  feature embedding,'' in \emph{ACM Multimedia}, 2014, pp. 675--678.

\bibitem{kalal12}
Z.~Kalal, K.~Mikolajczyk, and J.~Matas, ``Tracking-learning-detection,''
  \emph{TPAMI}, vol.~34, no.~7, pp. 1409--1422, 2012.

\bibitem{zhu2016robust}
G.~Zhu, F.~Porikli, and H.~Li, ``Robust visual tracking with deep convolutional
  neural network based object proposals on pets,'' in \emph{CVPR Workshops},
  2016, pp. 26--33.

\bibitem{zhang2016robust}
K.~Zhang, Q.~Liu, Y.~Wu, and M.-H. Yang, ``Robust visual tracking via
  convolutional networks without training,'' \emph{TIP}, vol.~25, no.~4, pp.
  1779--1792, 2016.

\bibitem{gao2014transfer}
J.~Gao, H.~Ling, W.~Hu, and J.~Xing, ``Transfer learning based visual tracking
  with gaussian processes regression,'' in \emph{ECCV}, 2014, pp. 188--203.

\end{thebibliography}
\end{document}